\documentclass[final]{cvpr}

\usepackage{cite}
\usepackage{float}
\usepackage{times}
\usepackage{epsfig}
\usepackage{graphicx}
\usepackage{amsmath}
\usepackage{amssymb}
\usepackage{booktabs}
\usepackage{mmstyles}
\usepackage{multirow}
\usepackage{caption}
\usepackage{rotating}

\usepackage{array}
\makeatletter
\def\hlinew#1{%
  \noalign{\ifnum0=`}\fi\hrule \@height #1 \futurelet
   \reserved@a\@xhline}
\makeatother

\newcommand{\printfnsymbol}[1]{%
  \textsuperscript{\@fnsymbol{#1}}%
}

\usepackage{makecell}

\newcolumntype{P}[1]{>{\raggedright\arraybackslash}p{#1}}
\newcolumntype{M}[1]{>{\centering\arraybackslash}m{#1}}

\usepackage[pagebackref=true,breaklinks=true,colorlinks,bookmarks=false]{hyperref}

\def\cvprPaperID{1619} 
\def\confYear{CVPR 2021}

\newcommand{\name}[1]{ForgeryNet}

\begin{document}

\title{ForgeryNet: A Versatile Benchmark for Comprehensive Forgery Analysis}

\author{Yinan He$^{1,2}$\thanks{} \quad Bei Gan$^{1,3*}$ \quad Siyu Chen$^{1,3*}$ \quad Yichun Zhou$^{1,4*}$ \\
Guojun Yin$^{1,3}$ \quad Luchuan Song$^{5}$\thanks{} \quad Lu Sheng$^{4}$ \quad Jing Shao$^{1,3}$\thanks{} \quad Ziwei Liu$^{6}$ \\ 
$^{1}$SenseTime Research \quad $^{2}$Beijing University of Posts and Telecommunications \\ $^{3}$Shanghai AI Laboratory             \quad $^{4}$College of Software, Beihang University \\ $^{5}$University of Science and Technology of China \quad
$^{6}$S-Lab, Nanyang Technological University \\
{\tt\small \{heyinan, ganbei, chensiyu, yinguojun, shaojing\}@sensetime.com}\\
{\tt\small \{buaazyc, lsheng\}@buaa.edu.cn} \quad 
{\tt\small slc0826@mail.ustc.edu.cn} \quad
{\tt\small ziwei.liu@ntu.edu.sg}
}

\vspace{-1cm}
\twocolumn[{%
\maketitle
\vspace{-20pt}
\begin{figure}[H]
    \hsize=\textwidth
    \centering
    \includegraphics[width=2.1\linewidth]{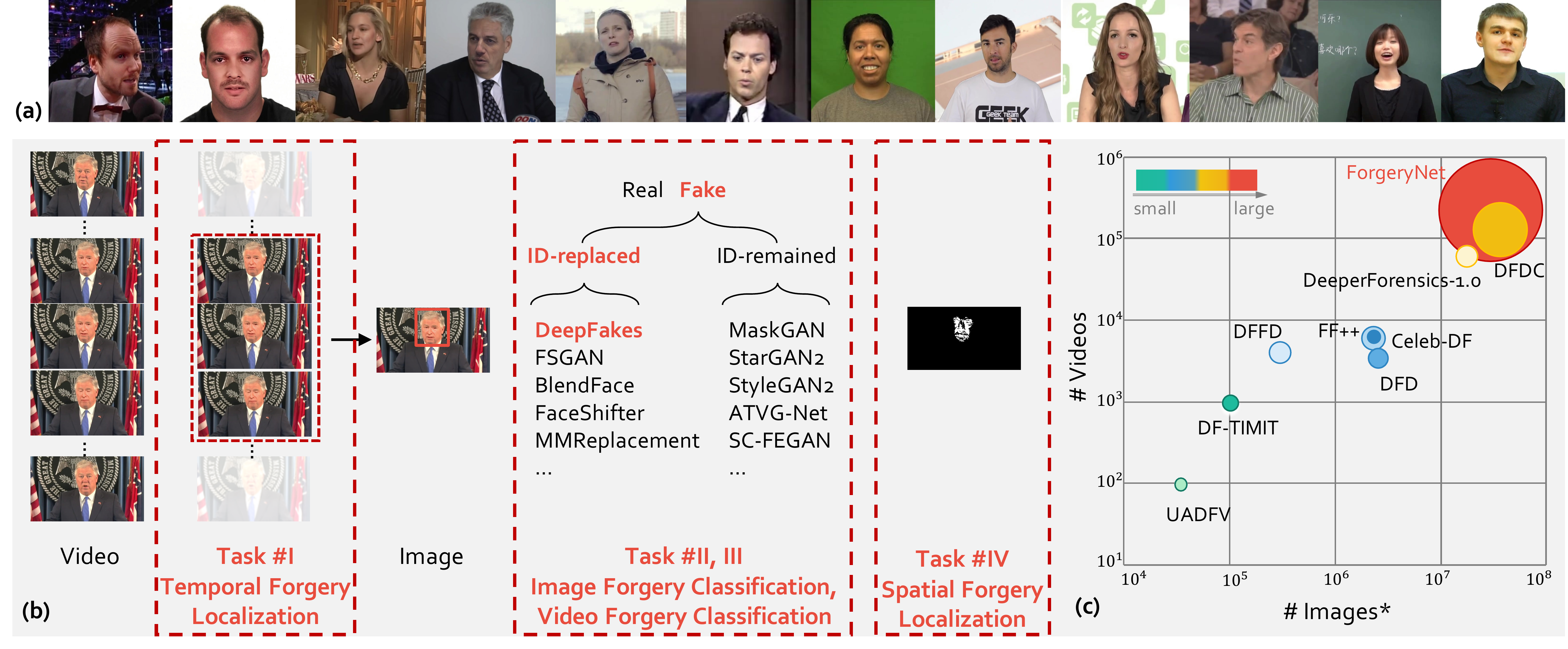}
    \vspace{-0.8cm}
    \captionof{figure}{ForgeryNet is a new mega-scale face forgery dataset with comprehensive annotations and four forgery analysis tasks. It contains thousands of subjects, various manipulation methods and diverse re-rendering processes. In (a), can you distinguish which images are forged?}
    \label{fig:fig1}
\end{figure}
}]

{
  \renewcommand{\thefootnote}%
    {\fnsymbol{footnote}}
  \footnotetext[1]{Equal contribution.}
  \footnotetext[2]{Work done during an internship at SenseTime Research.}
  \footnotetext[3]{Corresponding author.}
  \footnotetext[4]{\href{https://yinanhe.github.io/projects/forgerynet.html}{https://yinanhe.github.io/projects/forgerynet.html} }
}

\begin{abstract}
    \vspace{-0.3cm}
    The rapid progress of photorealistic synthesis techniques have reached at a critical point where the boundary between real and manipulated images starts to blur\footnotetext{\rotatebox[]{180}{ight): fake, fake, fake, fake, fake, fake, fake, real, fake, real, fake, real}  \rotatebox[]{180}{The label of images in Fig.~\ref{fig:fig1}(a)(from left to r-} }. Thus, benchmarking and advancing digital forgery analysis have become a pressing issue. However, existing face forgery datasets either have limited diversity or only support coarse-grained analysis. 
    
    \vspace{-0.1cm}
    To counter this emerging threat, we construct the \name{} dataset, an extremely large face forgery dataset with unified annotations in image- and video-level data across four tasks:
    1) \textbf{Image Forgery Classification}, including two-way (real / fake), three-way (real / fake with identity-replaced forgery approaches / fake with identity-remained forgery approaches), and $n$-way (real and $15$ respective forgery approaches) classification.
    2) \textbf{Spatial Forgery Localization}, which segments the manipulated area of fake images compared to their corresponding real images.
    3) \textbf{Video Forgery Classification}, which re-defines the video-level forgery classification with manipulated frames in random positions. This task is important because attackers in real world are free to manipulate any target frame.
    and 4) \textbf{Temporal Forgery Localization}, to localize the temporal segments which are manipulated.
    \name{} is by far the largest publicly available deep face forgery dataset in terms of data-scale (2.9 million images, 221,247 videos), manipulations (7 image-level approaches, 8 video-level approaches), perturbations (36 independent and more mixed perturbations) and annotations (6.3 million classification labels, 2.9 million manipulated area annotations and 221,247 temporal forgery segment labels). We perform extensive benchmarking and studies of existing face forensics methods and obtain several valuable observations. We hope that the scale, quality, and variety of our \name{} dataset will foster further research and innovation in the area of face forgery classification, as well as spatial and temporal forgery localization \etc.
    
    \end{abstract}
\vspace{-0.2cm}

\section{Introduction}
\label{sec:intro}

Photorealistic facial forgery technologies, especially recent deep learning driven approaches~\cite{petrov2020deepfacelab,li2019faceshifter,fried2019text}, give rise to widespread social concerns on potential malicious abuse of these techniques to eye-cheatingly forge media (\ie, images and videos, \etc) of human faces.
%
Therefore, it 
is of vital importance to develop reliable methods for face forgery analysis\footnote{In this paper, the definition of the term ``face forgery'' refers to an image or a video containing modified identity, expressions or attribute(s) with a learning-based approach, distinguished with 1) a so-called ``CheapFakes''~\cite{paris2019deepfakes} that are created with off-the-shelf softwares without learnable components and 2) ``DeepFakes'' that only refer to manipulations with swapped identities~\cite{dolhansky2020deepfake}.}, so as to distinguish \emph{whether} and \emph{where} an image or video is manipulated.

Most recent progress about face forgery analysis are sparked by gathering of face forgery detection datasets~\cite{rossler2019faceforensics++,dolhansky2020deepfake} and early attempts of profiling intrinsic characteristics within the forgery images.
However, performances on most datasets have already saturated (\ie~over $99\%$ accuracy~\cite{guarnera2020deepfake,yu2018attributing,marra2019incremental,hulzebosch2020detecting}) due to their limited scales (\eg~number of images/videos and subject identities) and limited diversity (\eg~forgery approaches, scenarios, realistic perturbations, \etc).
Moreover, in practical applications, it is often required to detect forged faces by locating tampered areas in an image and/or manipulated segments in an untrimmed video, rather than merely providing a binary label.

In this paper, we construct a new mega-scale dataset named \name{} with comprehensive annotations, consisting of two groups (\ie~image- and video-level) and four tasks for real-world digital forgery analysis.
We carefully benchmark existing forensics methods on \name{}.
Extensive experiments and in-depth analysis show that this larger and richer annotated dataset can boost the development of next-generation algorithms for forgery analysis.
Specifically, \name{} brings several unique advantages over existing datasets.

\noindent \textbf{\textit{(1) Wild Original Data.}} 
Most current datasets are captured under controlled conditions (\eg~environment, angles and lighting). We collect original data with diversified dimensions of angle, expression, identity, lighting, scenario and \etc from four datasets \cite{cao2014crema,livingstone2018ryerson,Chung18b,ephrat2018looking}. Note that all the original data have a \textit{Creative Commons Attribution} license that allows to share and adapt the material. 

\noindent \textbf{\textit{(2) Various Forgery Approaches.}} 
There are at most $8$ forgery approaches in all current datasets, while \name{} is manipulated by $15$ approaches, including face transfer, face swap, face reenactment and face editing. We choose approaches that span a variety of learning-based models, including encoder-decoder structure, generative adversarial network, graphics formation and RNN/LSTM (Fig.~\ref{fig:pipeline}).

\noindent \textbf{\textit{(3) Diverse Re-rendering Process.}} 
In the process of transmission and re-rendering, media data (image/video) always undergo compression, blurring and other operations, which may smooth the traces of forgery and bring more challenge for forgery detection. The \name{} dataset posts $36$ perturbations, such as optical distortion, multiplicative noise, random compression, blur, and \etc. As shown in Fig.~\ref{fig:fig1}(c), circle sizes refer to the number of forgery approaches with re-rendering process operations.

\noindent \textbf{\textit{(4) Rich Annotations and Comprehensive Tasks.}}
According to the real application scenario, we propose four tasks, as shown in Fig.~\ref{fig:fig1}(b): 
1) Image Forgery Classification, distinguishes whether an image is forgery or not and meanwhile tells its forgery type (\ie~manipulation approaches). We provide three types of annotations including two-way, three-way and $n$-way classification.
Both intra- and cross-forgery evaluations are set on three-way and $n$-way settings.
%
2) Spatial Forgery Localization, localizes manipulated areas of forgery images. Due to the fact that a forgery image may contain multiple faces and can be manipulated entirely or in part, it is more substantial to segment modified pixels in addition to only telling that it is forged.
3) Video Forgery Classification, similar to image-level classification, contains three types of annotations. Note that different from existing forgery video datasets, we construct our video dataset with untrimmed videos, each of which has part of the frames manipulated, considering the fact that forgery videos in real world are often manipulated on a certain subject and some key frames.
4) Temporal Forgery Localization, localizes the temporal segments which are manipulated. This is a new task for forgery analysis. Together with Video Forgery Classification and Spatial Forgery Localization, it provides comprehensive spatio-temporal forgery annotations.

\vspace{-0.2cm}
\section{Related Works}
\label{sec:related_works}

\begin{table*}[]
\centering
\caption{Comparison of various face forgery datasets. ForgeryNet surpasses any other dataset both in scale and diversity. It provides both video- and image-level data. The forgery data are constructed by $15$ manipulation approaches within $4$ categories. We also employ $36$ types of perturbations from $4$ kinds of distortions for post-processing.}
\label{tab:dataset-comparison-image}
\vspace{-0.3cm}
\footnotesize
\begin{tabular}{l|rrrrrrrrr}
\hlinew{1.1pt}
\multicolumn{1}{c|}{\multirow{2}{*}{Dataset}} & \multicolumn{2}{c}{Video Clips} & \multicolumn{2}{c}{Still images} & \multirow{2}{*}{Approaches} & \multirow{2}{*}{Subjects} & \multirow{2}{*}{\begin{tabular}[c]{@{}c@{}}Uniq. \\ \multicolumn{1}{c}{Perturb.}\end{tabular}} & \multirow{2}{*}{\begin{tabular}[c]{@{}c@{}}Mix \\ \multicolumn{1}{c}{Perturb.}\end{tabular}} & \multirow{2}{*}{Annotations} \\
\multicolumn{1}{c|}{} & \multicolumn{1}{c}{Real} & \multicolumn{1}{c}{Fake} & \multicolumn{1}{c}{Real} & \multicolumn{1}{c}{Fake} &  &  &  &  &  \\ \hline\hline
UADFV\cite{yang2019exposing} & 49 & 49 & 241 & 252 & 1 & 49 & - & $\times$ & 591 \\
DF-TIMIT\cite{korshunov2018deepfakes} & 320 & 640 & - & - & 2 & 43 & - & $\times$ & 1,600 \\
Deep Fake Detection\cite{GoogleDFD} & 363 & 3,068 & - & - & 5 & 28 & - & $\times$ & 3,431 \\
Celeb-DF\cite{li2020celeb} & 590 & 5,639 & - & - & 1 & 59 & - & $\times$ & 6,229 \\
SwapMe and FaceSwap\cite{zhou2017two} & - & - & 4,600 & 2,010 & 2 & - & - & $\times$ & 6,610 \\
DFFD\cite{dang2020detection} & 1,000 & 3,000 & 58,703 & 240,336 & 7 & - & - & $\times$ & 8,000 \\
FaceForensics++\cite{rossler2019faceforensics++} & 1,000 & 5,000 & - & - & 5 & - & 2 & $\times$ & 11,000 \\
DeeperForensics-1.0\cite{jiang2020deeperforensics} & 50,000 & 10,000 & - & - & 1 & 100 & 7 & \checkmark & 60,000 \\
DFDC\cite{dolhansky2020deepfake} & 23,564 & 104,500 & - & - & 8 & 960 & 19 & $\times$ & 128,064 \\
\textbf{ForgeryNet (Ours)} & \textbf{99,630} & \textbf{121,617} & \textbf{1,438,201} & \textbf{1,457,861} & \textbf{15} & \textbf{5400+}  & \textbf{36} & \textbf{\checkmark} & \textbf{9,393,574} \\ 
\hlinew{1.1pt}
\end{tabular}
\vspace{-0.3cm}
\end{table*}


Due to the urgency in detecting face manipulation, many efforts have been devoted to creating face forgery detection datasets. Previous datasets can be grouped down into three generations. Their statistical information is listed in Tab.~\ref{tab:dataset-comparison-image}.

\noindent\textbf{The first generation} consists of datasets such as DF-TIMIT~\cite{korshunov2018deepfakes}, UADFV~\cite{yang2019exposing}, SwapMe and FaceSwap~\cite{zhou2017two}. 
DF-TIMIT manually selects 16 pairs of appearance-similar people from the publicly available VidTIMIT database, and generates 640 videos with faces swapped. 
UADFV contains 98 videos, \ie~49 real videos from YouTube and 49 fake ones generated by FakeAPP~\cite{Fakeapp}. 
SwapMe and FaceSwap choose two face swapping Apps~\cite{Faceswap2018,Swapme} to create 2010 forgery images in total on 1005 original real images.

\noindent\textbf{The second generation} includes Google DeepFake Detection dataset~\cite{GoogleDFD} with 3,068 forgery videos by five publicly available manipulation approaches, and Celeb-DF~\cite{li2020celeb} containing 590 YouTube real videos mostly from celebrities and 5,639 manipulated video clips.
FaceForensics++~\cite{rossler2019faceforensics++} consists of 4000 fake videos manipulated by four approaches (\ie~DeepFakes, Face2Face, FaceSwap and NeuralTextures), and 1000 real videos from YouTube. 
The data scale and quality of the second generation have been improved. However, these datasets still lack diversity in forgery approaches and task annotations, and are not well-suited for challenges encountered in real world.

\noindent\textbf{The third generation} datasets are the most recent face forgery datasets, \ie~DeeperForensics-1.0~\cite{jiang2020deeperforensics}, DFDC~\cite{dolhansky2020deepfake}, and DFFD~\cite{dang2020detection} which contains tens of thousands of videos and tens of millions of frames. 
DeeperForensics-1.0 consists of 60,000 videos for real-world face forgery detection. 
DFDC contains over 100,000 clips sourced from 960 paid actors, produced with several face replacement forgery approaches including learnable and non-learnable approaches. 
In a practical application, in addition to classification, it is necessary to locate the manipulated areas or segments in an image or an untrimmed video. A few datasets have taken these tasks into consideration. DFFD provides annotations of spatial forgery at the first time, yet it only presents binary masks without manipulation density. 

\vspace{-0.2cm}
\begin{figure}[t]
\begin{center}
\includegraphics[width=1\linewidth]{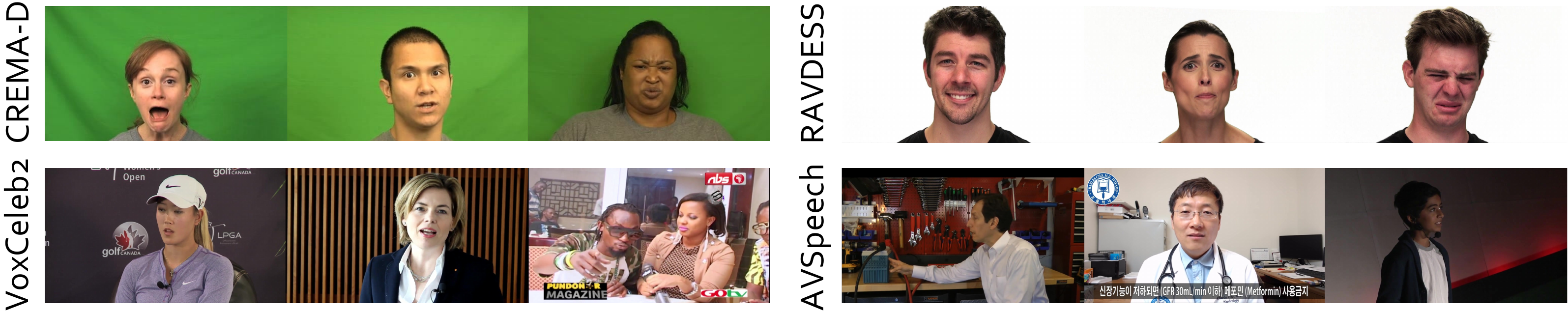}
\vspace{-1cm}
\end{center}
  \caption{Representative examples of original data collected from four face datasets respectively.}
\label{fig:sourcedata}
\vspace{-0.5cm}
\end{figure}

\section{\name{} Construction }
\label{sec:data_construction}

\begin{figure}[t]
   \begin{center}
   \includegraphics[width=0.9\linewidth]{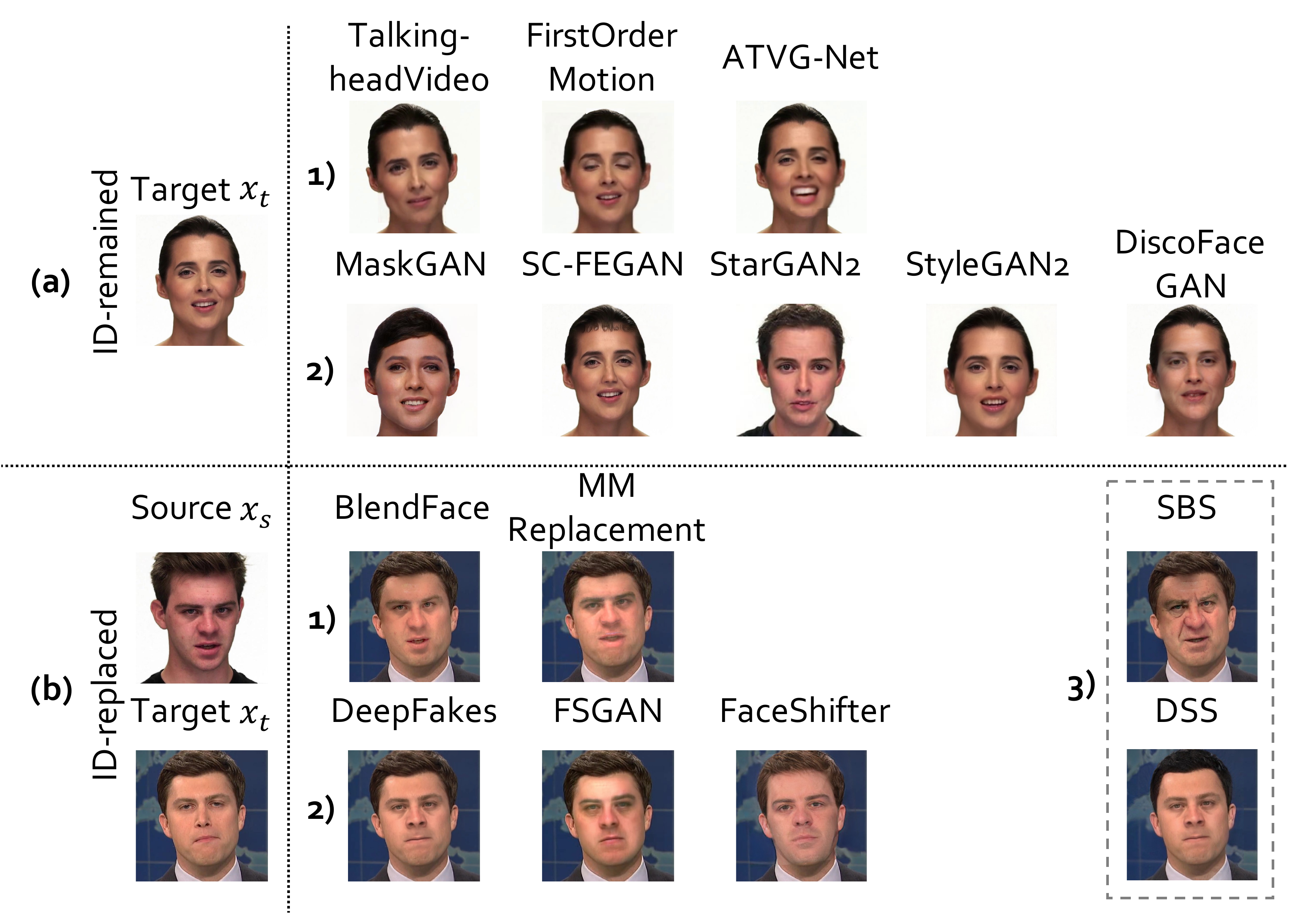}
   \vspace{-0.8cm}
   \end{center}
      \caption{Sampled forgeries in our ForgeryNet. (a) Identity-remained forgery approaches: 1) \textit{Face reenactment}, 2) \textit{Face editing}.
      (b) Identity-replaced forgery approaches: 1) \textit{Face transfer}, 2) \textit{Face swap}, 3) \textit{Face stacked manipulation}.}
      \vspace{-0.5cm}
   \label{fig:source_and_target}
\end{figure}

\begin{figure*}[t]
   \begin{center}
   \includegraphics[width=1.0\linewidth]{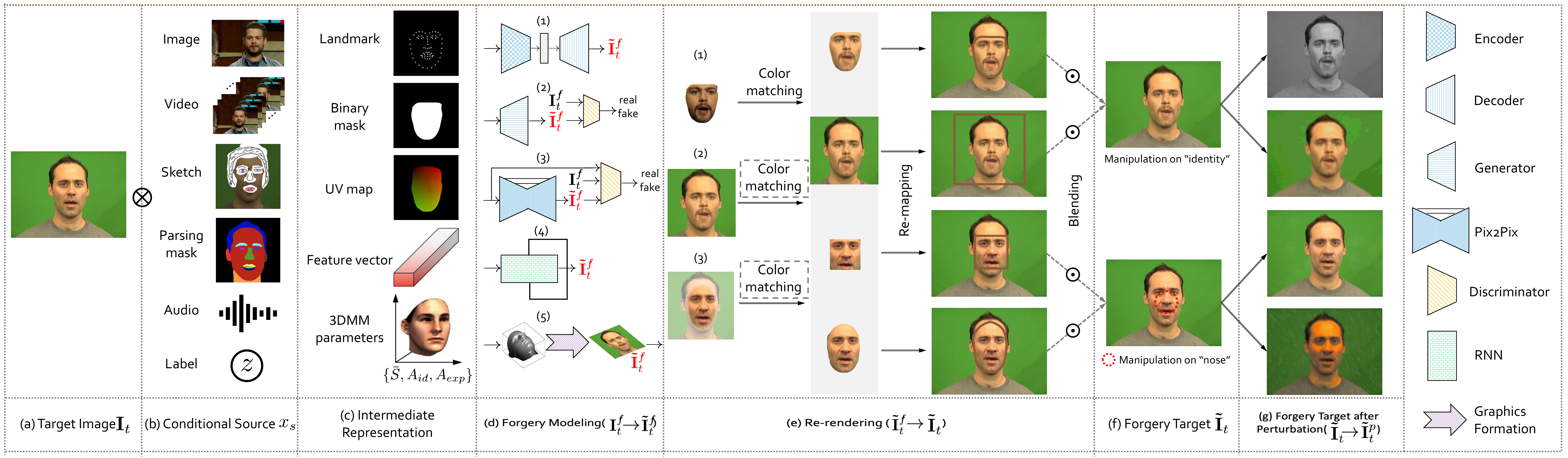}
   \end{center}
      \vspace{-0.5cm}
      \caption{
         Pipeline of face forgery approaches. (a)-(c) Representation preparation: target image $\mathbf{I}_t$, conditional source $x_s$ and their intermediate representations. (d) Forgery models produce a forged target face $\mathbf{\tilde{I}_t^f}$ by processing the representations. (e)-(f) Re-render $\mathbf{\tilde{I}_t^f}$ to full image $\mathbf{I}_t$ and get the forgery image $\mathbf{\tilde{I}_t}$. (g) Apply perturbations to $\mathbf{\tilde{I}_t}$ to obtain final forgery data.}
      \vspace{-0.5cm}
   \label{fig:pipeline}
\end{figure*}
Most of existing public face forgery datasets \cite{yang2019exposing,korshunov2018deepfakes,GoogleDFD,korshunov2018deepfakes,li2020celeb,zhou2017two,rossler2019faceforensics++,dang2020detection,jiang2020deeperforensics,dolhansky2020deepfake} contain only single or no more than 10 specific manipulation approaches, and even the largest one~\cite{dolhansky2020deepfake} only operates $8$ manipulations with $19$ perturbations on $960$ subjects. Moreover, these datasets take forgery analysis solely as a classification task. 
On the contrary, our proposed \name{} dataset provides $15$ manipulation approaches with more than $36$ mix-perturbations on over $5,400$\footnote{Some original datasets do not provide the identity annotation.}  subjects, and defines four tasks (\ie~image and video classification, spatial and temporal localization) with a total of $9.4$M annotations. 
Our whole dataset consists of two subsets: \textit{Image-forgery} set provides over $2.9$M still images and \textit{Video-forgery} set has more than $220$k video clips. These two subsets have their real data respectively randomly selected from the original data, and $15$ forgery approaches are applied to image-forgery construction while $8$ of them also generate the video-forgery data\footnote{There are $7$ forgery approaches that are only suitable for generating images.}.
We compare our \name{} with other publicly available datasets in Tab.~\ref{tab:dataset-comparison-image}. Over all the comparison items listed in the table, our dataset surpasses the rest both in scale and diversity.

\subsection{Original Data Collection}
\label{subsec:Original_Data_Collection}

\noindent\textbf{Source of Original Data.} Four face datasets, CREMA-D~\cite{cao2014crema}, RAVDESS~\cite{livingstone2018ryerson}, VoxCeleb2~\cite{Chung18b} and AVSpeech~\cite{ephrat2018looking}, are chosen as the original data to boost the diversity in dimensions of face identity, angle, expression, scenarios \etc.
%
%

Note that CREMA-D is made available under the Open Database License, while others are released under a Creative Commons Attribution License.
The resolutions of these original data range from $240$p to $1080$p, and face yaw angles ranging from $-90$ to $90$ degrees are all covered. Representative examples are shown in Fig.~\ref{fig:sourcedata}.
%




%

\noindent\textbf{Preprocess Original Data.}
For further manipulation, we crop original videos into a controllable set of source videos with reasonable lengths.
Then we detect and select faces for manipulation and obtain their face attribute labels.
\subsection{Forgery Approach}
\label{subsec:Forgery_Approach}

To guarantee the diversity of forgery approaches in the proposed \name{}, we introduce $15$ face forgery approaches\footnote{Detailed description of the forgery approaches is provided in the appendix.}~\cite{Siarohin_2019_NeurIPS,chen2019hierarchical,fried2019text , choi2020stargan,Karras2019stylegan2, CelebAMask-HQ, Jo_2019_ICCV,deng2020disentangled ,nirkin2019fsgan ,petrov2020deepfacelab , li2019faceshifter}. 
They are selected according to perspectives of modeling types, conditional sources, forgery effects and functions.
We denote $x_t$ as the \textit{target} subject to be manipulated while the \textit{source} $x_s$ is regarded as the conditional media driving the \textit{target} to change either identity or attributes, or even both.

\vspace{-0.3cm}
\subsubsection{Forgery Category}
\label{subsubsec:forgery_category}
\vspace{-0.2cm}
According to the visual effects of facial manipulation, we divide the forgery approaches into two categories, \ie~\textit{Identity-remained} and \textit{Identity-replaced}.
Sampled forgeries in Fig.~\ref{fig:source_and_target} illustrate these categories and their sub-types.

\vspace{0.1cm}
\noindent \textbf{Identity-remained Forgery Approach} in Fig.~\ref{fig:source_and_target}(a)
remains the identity of $x_t$ and the identity-agnostic content like expression, mouth, hair and pose of $x_t$ are changed, driven by $x_s$. We adopt eight approaches and divide them into two sub-types:
1) \textit{Face reenactment} on $x_t(i,a)$ preserves its identity but has its \textit{intrinsic} attributes like pose, mouth and expression manipulated by conditional source $x_s$ and forms $x_t(i,\tilde{a}^s)$, where $i$ refers to identity and $a$ denotes attribute(s). Alternatively, with 2) \textit{Face editing} on $x_t(i,a)$ has its \textit{external} attributes altered, such as facial hair, age, gender and ethnicity, to obtain $x_t(i,\hat{a}^s)$. We also include multiple attribute manipulation with two editing approaches, \eg both hair and eyebrow are manipulated as shown with the first example in Fig.~\ref{fig:source_and_target}(a-2).

\vspace{0.1cm}
\noindent \textbf{Identity-replaced Forgery Approach} in Fig.~\ref{fig:source_and_target}(b)
replaces the content of $x_t$ with that of $x_s$ preserving the identity of $s$. Seven approaches are divided into three sub-types as follows.
1) \textit{Face transfer} transfers both identity-aware and identity-agnostic content (\eg~expression and pose) from $x_s$ to $x_t$, resulting in $x_t(\tilde{i}^s,\tilde{a}^s)$. 
2) \textit{Face swap} which produces $x_t(\tilde{i}^s,a)$ only swaps identity from the source $x_s$ to the target $x_t$, and the identity-agnostic content $a$ are preserved.
3) \textit{Face stacked manipulation} refers to a combination of both \textit{Identity-remained} and \textit{Identity-replaced} approaches. We propose two assembles\footnote{StarGAN2-BlendFace-Stack (SBS), DeepFakes-StarGAN2-Stack (DSS)}, \ie~$\langle\textit{editing}\rightarrow\textit{transfer}\rangle$ and $\langle\textit{swap}\rightarrow\textit{editing}\rangle$, where the former one transfers both the identity and attributes of the manipulated $x_s(i,\hat{a})$ to the target $x_t$ to obtain $x_t(\tilde{i}^s,\tilde{\hat{a}}^s)$ and the latter alters the external attributes of the swapped target $x_t(\tilde{i}^s,a)$ to get $x_t(\tilde{i}^s,\hat{a}^s)$.

\vspace{-0.3cm}
\subsubsection{Forgery Pipeline}

\vspace{-0.2cm}
Although there are a wild variety of architectures designed for the aforementioned approaches, most are created using variations or combinations of generative networks, encoder-decoder networks or graphics formation. We briefly summarize the forgery pipeline in Fig.~\ref{fig:pipeline}.

The target is always an image marked as $\mathbf{I}_t$, while there are various conditional source formats $x_s$, including image, image sequence, sketch map, parsing mask, audio, label, or even noise. 
We first detect the \textit{target} face $\mathbf{I}_t^f$, crop and align it, and then transform both the \textit{target} face as well as \textit{source} data to intermediate representations such as UV map, feature bank, 3DMM parameters and \etc. 

\vspace{0.1cm}
\noindent \textbf{Forgery Modeling.}
These representations are forwarded to the forgery models to obtain a forged target face $\mathbf{\tilde{I}}_t^f$. We include five architecture variants as, 
1) \textit{Encoder-Decoder}~\cite{faceswap}, 
2) \textit{Vanilla GAN}~\cite{shen2018facefeat}, 
3) \textit{Pix2Pix}~\cite{li2019faceshifter}, 
4) \textit{RNN/LSTM} \cite{chen2019hierarchical}, 
and 5) \textit{Graphics Formation}~\cite{egger20203d}.

\begin{figure}[t]
\begin{center}
\includegraphics[width=1.0\linewidth]{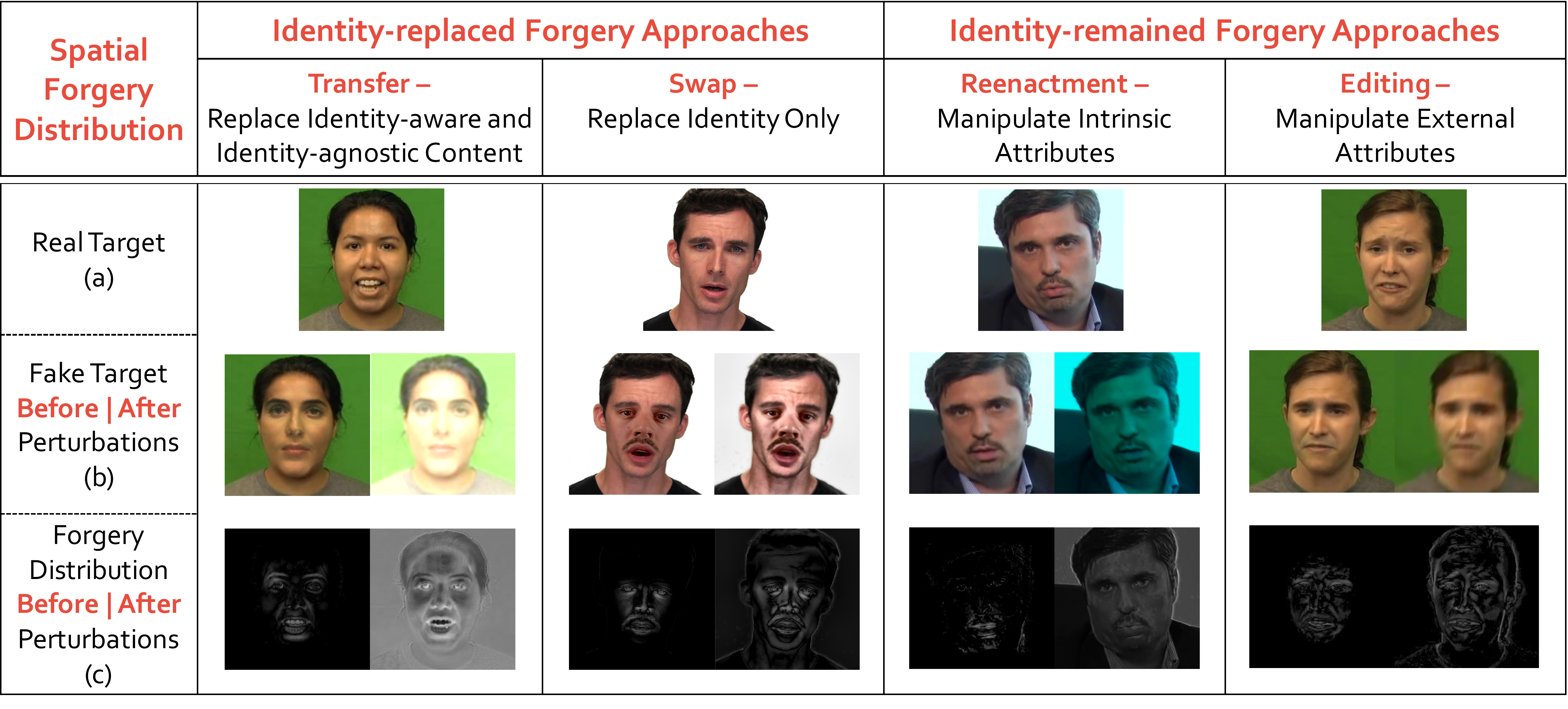}
\end{center}
  \vspace{-0.7cm}
  \caption{Annotations for Spatial Forgery Localization in \name{}. Examples of (a) real image, (b) forgery image, (c) corresponding spatial annotations.}
  \vspace{-0.3cm}
\label{fig:facemanipulationmask}
\end{figure}

\noindent \textbf{Re-rendering Process.}
To acquire the full forged target, the forged target face $\mathbf{\tilde{I}}_t^f$ is re-rendered back to the target full image $\mathbf{I}_t$ to obtain $\mathbf{\tilde{I}}_t$. 
In particular, according to different forgery procedures,
1) $\mathbf{\tilde{I}}_t^f$ can be a \textit{face mask}, shown in Fig.~\ref{fig:pipeline}(e-1), which contains the area from the eyebrows to the face chin.
2) $\mathbf{\tilde{I}}_t^f$ can also be a \textit{face bounding-box}, illustrated in Fig.~\ref{fig:pipeline}(e-2,3), which keeps the same bounding box as the original target face.
\noindent \textbf{Perturbation.} 
To better reflect real-world data distribution, we apply 36 types of perturbations to the forgery data $\mathbf{\tilde{I}}_t$. 
We follow common practices in visual quality assessment~\cite{shahid2014no} with distortions of compression, transmission, capture, color, \etc.


\subsection{ForgeryNet Annotation}
\label{subsec:dataset_annotation}

In contrast to most previous datasets, our \name{} is annotated comprehensively both in image- and video-level across four tasks.

\noindent \textbf{Image Forgery Classification.}
According to the forgery definition in Sec.~\ref{subsubsec:forgery_category}, given a forgery image, we provide three types of forgery labels, \ie labels for two-way (real / fake), three-way (real / fake with identity-replaced forgery approaches / fake with identity-remained forgery approaches), and $n$-way ($n=16$, real and $15$ respective forgery approaches) classification tasks respectively. These annotations make it possible to explore the correlation between different forgery meta-types or approaches.

\noindent \textbf{Spatial Forgery Localization.}
As shown in Fig.~\ref{fig:facemanipulationmask}, we take the forgery image $\mathbf{\tilde{I}}_t$ and the corresponding real image $\mathbf{I}_t$ to calculate their difference to obtain a
\textit{forgery distribution} $\mathbf{\tilde{I}}_t^d$. 
In this paper, we define the \textit{Spatial Forgery Localization} task as ``\textit{localizing the face area manipulated by deep forgery approaches}'', and thus the forgery distribution before perturbation $\mathbf{\tilde{I}}_t^d$ is taken as the ground-truth annotation.

\noindent \textbf{Video Forgery Classification \& Temporal Forgery Localization.}
Note that in contrast to all the existing datasets, we construct our video forgery dataset with untrimmed forgery videos $\mathbf{\tilde{V}}_t^{\prime}$, each of which splices real and manipulated frames together.
%
%
Same as image-forgery, \textit{Video Forgery Classification} also contains three types of class annotations. 
We also provide the annotations on locations of manipulated segments in the untrimmed forgery video and propose a new task, \ie~\textit{Temporal Forgery Localization}, to localize these forged segments. 
%
\begin{figure}[t]
\begin{center}
\includegraphics[width=1.0\linewidth]{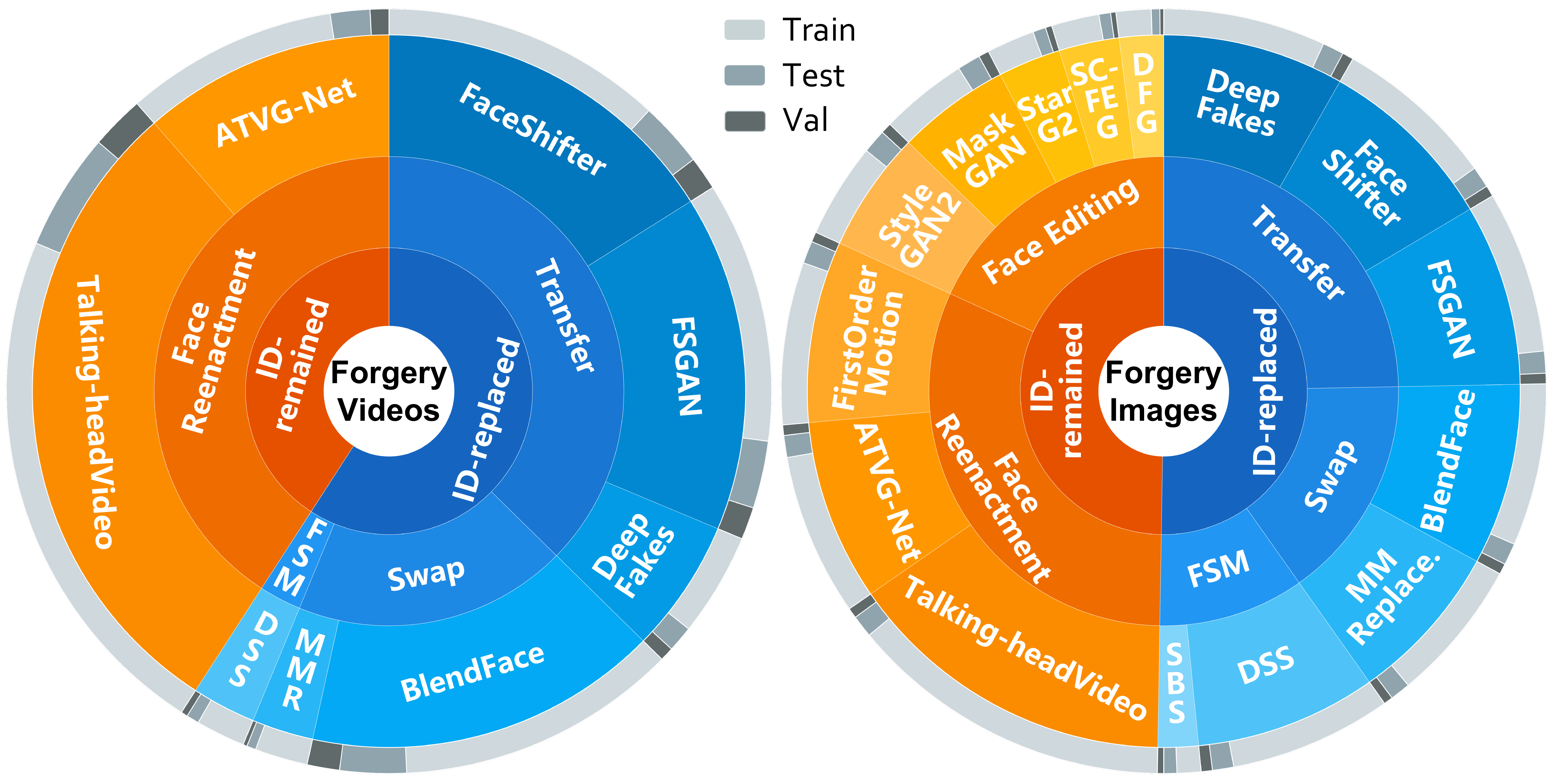}
\end{center}
   \vspace{-0.7cm}
   \caption{Illustration of image- and video-level sets. From the inside to the outside are categories of \textit{Identity-remained} and \textit{Identity-replaced}, corresponding sub-types, specific forgery approaches and the situation of data split. 
   }
   \vspace{-0.3cm}
\label{fig:videoAndImageNums}
\end{figure}


\section{ForgeryNet Settings}
\label{sec:benchmark}

On \name{}, we set up two benchmarks, image and video, with a series of tasks for face forgery analysis.
%

\noindent\textbf{Dataset Preparation.}
Both image- and video-level sets are split into training, validation and test subsets with a ratio close to 7:1:2. Forgery data distributions and catagories of the two sets are shown in Fig.~\ref{fig:videoAndImageNums}. Forgery data in each subset have identities matched with the corresponding real subset. The ratio of real to fake in each subset is close to 1:1.

\subsection{Image Benchmark Settings}
\label{subsubsec:Forgery Image Analysis}

\subsubsection{Image Forgery Classification}
\label{subsubsec:Image_Forgery_Classification}

In order to foster further researches on face forgery classification, we carefully design two protocols to evaluate forensics methods in this area.

\noindent\textbf{Protocol 1: Intra-forgery Evaluation.}
In intra-forgery evaluation, all the real and fake data in the training set are used to train models, and the validation set is used for evaluation.
%
This protocol has three variants, according to the definition in Sec.~\ref{subsec:dataset_annotation}, \ie~two-/three-/$n$-way classification.
\begin{table}[]
    \caption{\textbf{Image Forgery Classification (Protocol 1):} binary classification. We report accuracy and AUC scores of the compared forensics methods.}
    \vspace{-0.3cm}
    \footnotesize
    \centering
    \label{tab:image-class-table}
    \begin{tabular}{l|rrr}
       \hlinew{1.1pt} 
       Method  &  Param. & Acc & AUC \\ \hline\hline
       MobileNetV3 Small~\cite{howard2019searching}& 1.7M & 76.24 & 85.51 \\
       MobileNetV3 Large~\cite{howard2019searching} & 4.2M & 78.30 & 87.56 \\
       EfficientNet-B0~\cite{tan2019efficientnet}  & 4.0M & 79.86 & 89.31 \\
       ResNet-18~\cite{he2016deep} & 11.2M & 78.31 & 87.75 \\ 
       Xception~\cite{chollet2017xception}  & 20.8M & 80.78 & 90.12 \\ 
       ResNeSt-101~\cite{zhang2020resnest}  & 46.2M & 82.06 & 91.02 \\ \hline
       SAN19-patchwise~\cite{zhao2020exploring}  & 18.5M & 80.08 & 89.38 \\
       ELA-Xception~\cite{gunawan2017development}  & 20.8M & 73.77 & 82.69 \\
       SNRFilters-Xception~\cite{chen2017jpeg}  & 20.8M & 81.09 &90.52\\
       GramNet~\cite{liu2020global}  & 22.1M & 80.89 & 90.20 \\ 
       F$^3$-Net~\cite{qian2020thinking}  & 57.3M & 80.86 & 90.15 \\
       \hlinew{1.1pt}
    \end{tabular}
    \vspace{-0.3cm} 
 \end{table}

\noindent\textbf{Protocol 2: Cross-forgery Evaluation.}
To further evaluate the generalization ability of training with our data, we conduct cross-forgery evaluation by training the evaluated forensics method with one certain type of manipulation and testing it with others. The manipulation type can either be general (\eg \textit{identity-replaced}), or specific (\eg \textit{ATVG-Net}). Note that this protocol only involves binary classification.

\noindent\textbf{Metrics.}
For binary classification tasks, we evaluate with Accuracy (Acc) and the Area under ROC curve (AUC).
For three- and $n$-way class settings, we use Accuracy (Acc) and mean Average Precision (mAP) as evaluation metrics.
\vspace{-0.3cm}
\subsubsection{Spatial Forgery Localization}
\label{subsubsec:Image_Forgery_Localization}
\vspace{-0.2cm}
Compared with classification tasks, spatial forgery localization aims to specify manipulated regions.
Images along with forgery masks are used to train the localization model.
%
%

\noindent\textbf{Metrics.}
We utilize three metrics for evaluation: two variants of Intersection over Union (IoU) and L1 distance.
%

\subsection{Video Benchmark Settings}

\noindent\textbf{Video Forgery Classification. } 
Evaluation protocols for video forgery classification are generally similar to the ones designed for the image set, except that $n$=9 for $n$-class setting.
Metrics are the same as those for image classification.

\noindent\textbf{Temporal Forgery Localization. }
%
%
%
For each video, forensics methods to be evaluated are expected to provide temporal boundaries of forgery segments and the corresponding confidence values.
%
We follow metrics used in ActivityNet~\cite{ghanem2018activitynet} evaluation, and employ Interpolated Average Precision (AP) as well as Average Recall@$K$ (AR@$K$) for evaluating predicted segments with respect to the groundtruth ones.
%
\vspace{-0.2cm}
%
\begin{figure*}[t]
   \includegraphics[width=1.0\linewidth]{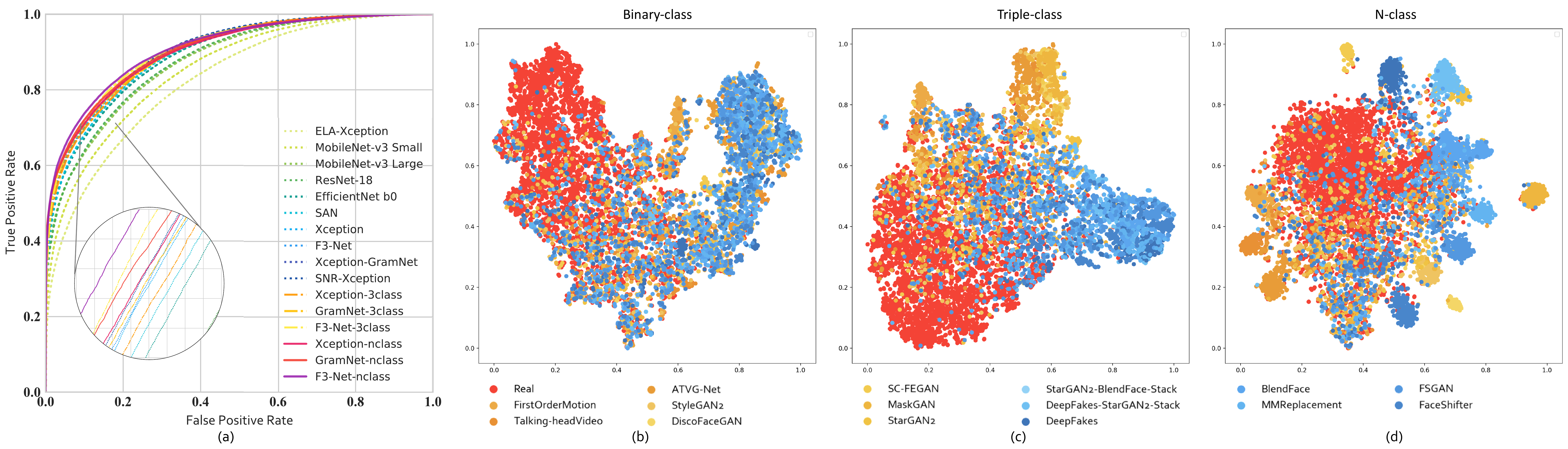}
   \vspace{-0.5cm}
      \caption{\textbf{Image Forgery Classification (Protocol 1):} (a)  We show the ROC curves of the compared methods under the setting of binary classification. (b)-(d) t-SNE feature visualization of the data manipulated by different forgery approaches, trained with binary, three-way and $n$-way classification respectively.}
    \vspace{-0.5cm}
   \label{fig:img_cls_rocand_multicls}
   \end{figure*}
%
\section{Image Forgery Analysis Benchmark}
\label{sec:image_benchmark}

\subsection{Image Forgery Classification}
\begin{figure}[t]
   \includegraphics[width=1.0\linewidth]{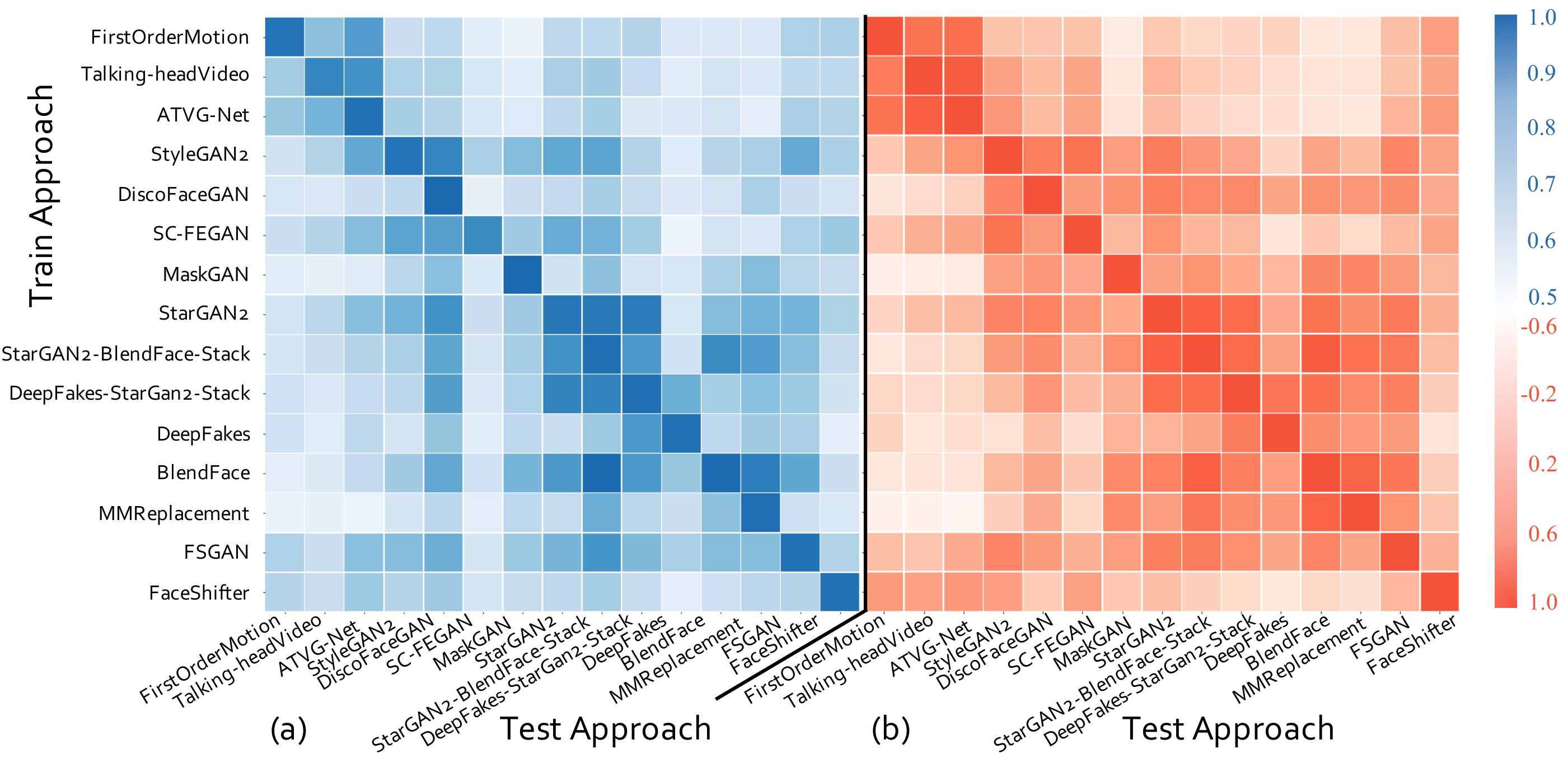}
   \vspace{-0.6cm}
      \caption{\textbf{Image Forgery Classification (Protocol 2):} 
      %
      (a) AUC score map, and (b) correlation map according to the AUC scores.
      X-axis denotes the tested forgery approach and  Y-axis denotes the forgery approach for training.
      }
      \vspace{-0.5cm}
      \label{fig:img_cross_method}
\end{figure}

\noindent\textbf{Protocol 1: Intra-forgery Evaluation.}
For comprehensive evaluation, we provide results of two-way class classification with several representative models of different sizes. 
%
Considering the trade-off between performance and efficiency, we use Xception~\cite{chollet2017xception} as the baseline model. ELA-Xception~\cite{gunawan2017development} and SNRFilters-Xception~\cite{chen2017jpeg} are two variants of Xception. Smaller models include MobileNetV3~\cite{howard2019searching}, EfficientNet-B0~\cite{tan2019efficientnet} and ResNet-18~\cite{he2016deep}. We select ResNeSt-101~\cite{zhang2020resnest} as the large model. We also experiment with recent state-of-the-art methods for face forgery detection, \ie F$^3$-Net~\cite{qian2020thinking} and GramNet~\cite{liu2020global}, as well as a fully-attentional network SAN19~\cite{zhao2020exploring}.
\begin{table}[]
   \caption{\textbf{Image Forgery Classification (Protocol 1):} multi-class settings and their mappings to binary classification. We report the accuracy, mAP and AUC scores. }
   \vspace{-0.3cm}
   \small
   \centering
   \label{tab:multi-class-and-cross-perturbation}
   \begin{tabular}{l|rrrr}
   \hlinew{1.1pt}
   \multirow{2}{*}{} & \multicolumn{2}{c}{3-way class} & \multicolumn{2}{c}{3$\rightarrow$2-way class}  \\
     & Acc. & mAP & Acc. & AUC \\ \hline\hline
   Xception  & 73.00 & 89.90& 80.17 & 89.92  \\
   GramNet & 73.30 & 90.00 & 80.75 & 90.13  \\
   F$^3$-Net & 74.45 & 90.41  & 81.75 & 90.63 \\ \hlinew{1.05pt}
   \multirow{2}{*}{} & \multicolumn{2}{c}{16-way class} & \multicolumn{2}{c}{16$\rightarrow$2-way class}  \\
    & Acc. & mAP  & Acc. & AUC \\ \hline\hline
   Xception & 58.81 & 93.16 & 81.00 & 90.53 \\
   GramNet  & 56.77 & 92.27& 80.83 & 90.25 \\
   F$^3$-Net  & 59.82 & 92.98& 81.88 & 90.91  \\ \hlinew{1.1pt}
   \end{tabular}
   \vspace{-0.1cm}
\end{table}

\begin{table}[]
   \caption{
   \textbf{Image Forgery Classification (Protocol 2):} binary classification. We report the accuracy and AUC scores.
   Forensics methods trained with ID-replaced forgery approaches have significant performance drops when tested on unseen ID-remained forgery approaches, and \emph{vice versa}.
   }
   \vspace{-0.3cm}
   \label{tab:image-cross-method}
   \small
   \centering
   \begin{tabular}{cl|cccc}
   \hlinew{1.1pt}
   \multicolumn{2}{c|}{\multirow{2}{*}{}} & \multicolumn{2}{c}{ID-replaced} & \multicolumn{2}{c}{ID-remained} \\
   \multicolumn{2}{c|}{} & Acc. & AUC & Acc. & AUC \\ \hline \hline
   \multirow{2}{*}{Xception} & ID-replaced & 84.13 & 92.80 & 64.62 & 74.86 \\
    & ID-remained & 67.28 & 75.83 & 81.17 & 90.71 \\
   \multirow{2}{*}{GramNet} & ID-replaced & 82.82 & 92.54 & 62.72 & 74.28 \\
    & ID-remained & 67.50 & 76.19 & 80.60 & 90.28 \\
  \multirow{2}{*}{F$^3$-Net} & ID-replaced & 83.84 & 92.73 & 64.33 & 73.82 \\
    & ID-remained & 68.44 & 77.24 & 81.18 & 90.29 \\ \hlinew{1.1pt}
   \end{tabular}
\end{table}

All experiments are conducted on face images cropped with face bounding boxes enlarged by $1.3\times$.
During training, we use several types of data augmentation to mimic distortions caused by compression and packet loss during transmission, so as to improve the generalization of developed models.

As presented in Tab.~\ref{tab:image-class-table}, we list binary classification metrics of all aforementioned forensics methods. 
We also show the corresponding ROC curves of these methods in Fig.~\ref{fig:img_cls_rocand_multicls}(a).
For three-way and $16$-way classification experiments, as shown in Tab.~\ref{tab:multi-class-and-cross-perturbation}, Acc scores show that classification becomes more difficult when the number of categories increases, yet the mAP metric indicates that the discrimination ability becomes higher instead. Moreover, after mapping back to binary classification, we can also observe slight performance boosts on F$^3$-Net compared to training results with only binary labels. This suggests that more auxiliary information potentially makes the forensics model more discriminative.
%
%
%
%

%
%


\noindent\textbf{Protocol 2: Cross-forgery Evaluation.}
For this protocol, we show the generalization ability of forensics methods across forgery approaches. Tab.~\ref{tab:image-cross-method} lists the results of models trained on \textit{ID-replaced} but evaluated on \textit{ID-remained}, and \emph{vice versa}. The more exhaustive cross-forgery setting with 15 specific forgery approaches is also evaluated and shown in Fig.~\ref{fig:img_cross_method}.
%
We observe from these results that intra-forgery testing naturally performs the best.
%
%
%
From Fig.~\ref{fig:img_cross_method}(a), we can also see that training on \textit{ATVG-Net, StyleGAN2} or \textit{BlendFace} gives the best generalization performance on average. On the other hand, \textit{DiscoFaceGAN} is the most generalizable forgery approach, while \textit{SC-FEGAN} is the most difficult approach to generalize to.
There is another interesting finding that forgery approaches with stronger similarity tend to induce better cross-forgery performance. For example, \textit{DiscoFaceGAN} is a \textit{StyleGAN}-based approach, thus training on the latter approach produces favorable results on the former.
Similarly, \textit{StarGAN2} and the two face stack manipulations which both involve \textit{StarGAN2} generalize well to each other.
%
%
%
In addition, as shown in Fig.~\ref{fig:img_cross_method}(b), forgery approaches belonging to the same meta-category usually have higher correlations mutually.
For example, for meta-category \textit{Face reenactment}, if a forensics method can obtain good performance on \textit{ATVG-Net}, it may also work for \textit{FirstOrderMotion} and \textit{Talking-headVideo}.

\begin{table}[t]
    \caption{
    \textbf{Spatial Forgery Localization.}
    We compare results with three metrics, \ie, IOU, IOU$_\text{diff}$ and L1 distance.
    }
    \vspace{-0.3cm}
    \scriptsize
    \centering
    \label{tab:spacial-localize}
    \begin{tabular}{l|rr|rrr|r}
    \hlinew{1.1pt}    
    \multirow{2}{*}{Method} & \multicolumn{2}{c|}{IoU} & \multicolumn{3}{c|}{IoU$_{\text{diff}}$} & \multirow{2}{*}{Loss$_{l1}$} \\
       & 0.1 & 0.2 & 0.01 & 0.05 & 0.1 & \\ \hline \hline
    Xception+Reg. & 89.55 & 93.70 & 67.57 & 83.25 & 89.22 & 0.0131 \\
    Xeption+Unet~\cite{ronneberger2015u} & 95.99 & 98.76 & 79.71 & 92.70 & 97.13 & 0.0134\\
    HRNet~\cite{wang2020deep} & 96.27 & 98.78 & 88.73 & 92.99 & 96.27 & 0.0114\\ \hlinew{1.1pt}
    \end{tabular}
    \vspace{-0.4cm}
\end{table}
\begin{figure}[t] 
   \includegraphics[width=1.0\linewidth]{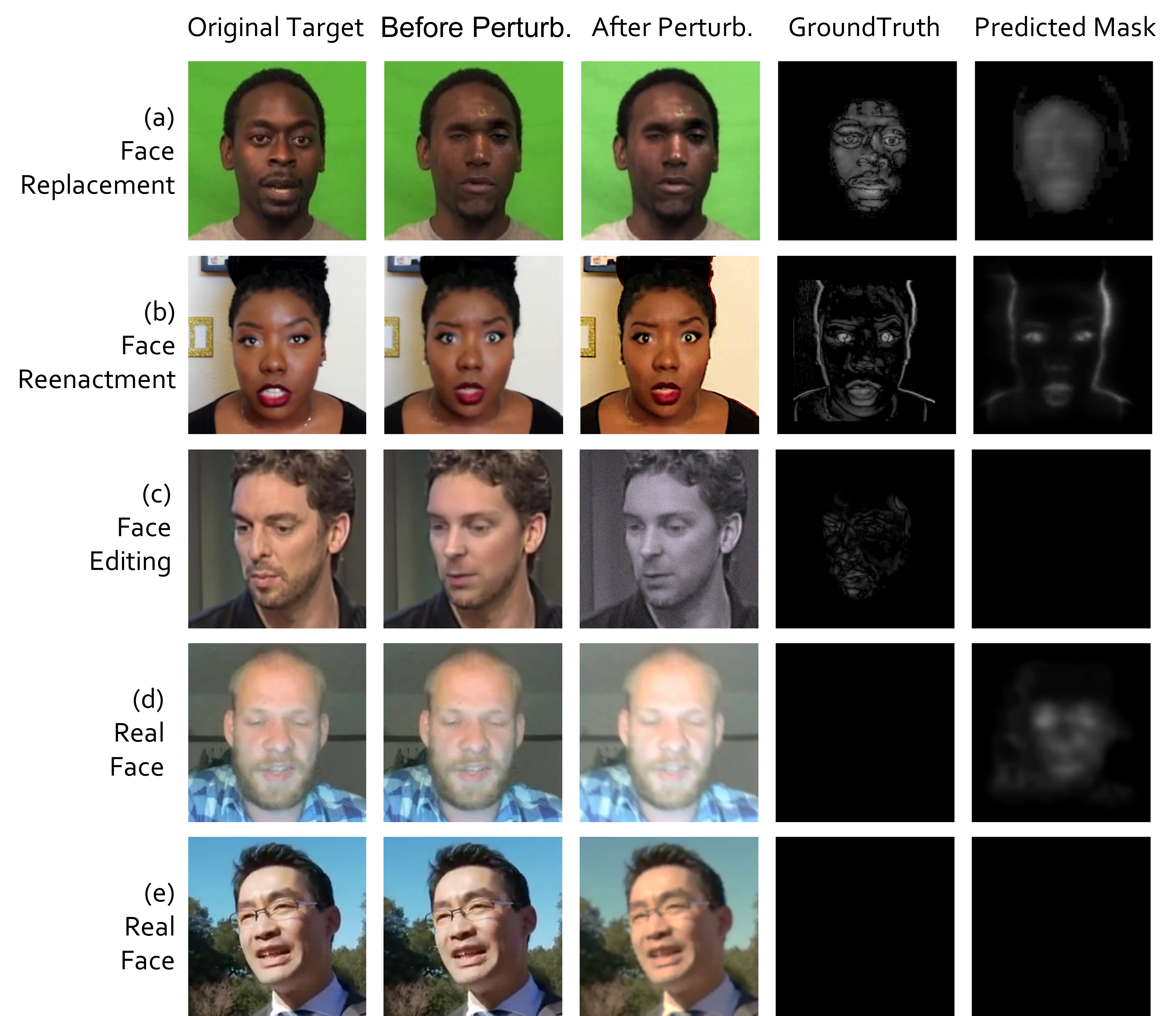}
   \caption{\textbf{Spatial Forgery Localization.} Examples of predicted manipulation masks by HRNet.
   %
   }
\label{fig:spaciallocalize_fake}
\vspace{-0.4cm}
\end{figure}

\subsection{Spatial Forgery Localization}
\label{subsec:forgery_spatial_localization}
We evaluate pixel regression and other two segmentation methods for the spatial localization task. UNet~\cite{ronneberger2015u} is a popular segmentation architecture, which has been widely used. For comparison, we also adopt HRNet~\cite{wang2020deep} because of its superior performance on other datasets. 

In Tab.~\ref{tab:spacial-localize}, HRNet outperforms other methods. Especially in terms of $\text{IoU}_\text{diff}$ with threshold $0.01$, HRNet surpasses other methods by more than $10\%$.
%
We also present predicted manipulation maps for several test samples. In Fig.~\ref{fig:spaciallocalize_fake}(c), the slight beard change is hard to detect, while in Fig.~\ref{fig:spaciallocalize_fake}(d), a real image is misjudged as manipulated.



\vspace{-0.2cm}
\section{Video Forgery Analysis Benchmark}
\label{sec:video_benchmark}

\begin{table}[t]
   \caption{\textbf{Video Forgery Classification (Protocol 1):} binary classificaiton. We report accuracy and AUC scores under two crop strategies. Video-level classification has better results than the image-level setting.}
   \vspace{-0.3cm}
    \scriptsize
    \centering
    \label{tab:Video cls Results}
    \begin{tabular}{l|r|rr|rr}
    \hlinew{1.1pt}
      &  & \multicolumn{2}{c|}{Single-crop} & \multicolumn{2}{c}{Multi-crop} \\
    Method  & Parameters & \multicolumn{1}{c}{Acc} & \multicolumn{1}{c|}{AUC} & \multicolumn{1}{c}{Acc} & \multicolumn{1}{c}{AUC} \\ 
    \hline
    \hline
    X3D-M~\cite{feichtenhofer2020x3d} & 2.9M & 87.93 & 93.75 & 88.97 & 96.99 \\
    Slow-only~\cite{feichtenhofer2019slowfast}& 31.6M & 86.76 & 92.64 & 87.37 & 95.96 \\
    TSM~\cite{lin2019tsm}  & 23.5M & 88.04 & 93.05 & 89.11 & 96.25 \\
    SlowFast~\cite{feichtenhofer2019slowfast}  & 33.6M & 88.78 & 93.88 & 89.92 & 97.28 \\ \hlinew{1.1pt}
    \end{tabular}
\end{table}

\begin{table}[t]
   \caption{\textbf{Video Forgery Classification (Protocol 1):} multi-class settings and their mappings to binary classification. We report the accuracy, mAP and AUC scores.}
   \vspace{-0.3cm}
   \centering
   \footnotesize
   \label{tab:video_multiclass}
   \begin{tabular}{l|rrrr}
  \hlinew{1.1pt}
   \multirow{2}{*}{Method} &  \multicolumn{2}{c}{3-way class} & \multicolumn{2}{c}{3$\rightarrow$2-way class} \\
    & Acc. & mAP & Acc. & AUC  \\ \hline\hline
   X3D-M~\cite{feichtenhofer2020x3d} & 84.00 & 94.55 & 87.69 & 93.78 \\
   SlowFast~\cite{feichtenhofer2019slowfast} & 85.73 & 94.89 & 89.11 & 94.37 \\ \hlinew{1.05pt}
   \multirow{2}{*}{} &  \multicolumn{2}{c}{9-way class} & \multicolumn{2}{c}{9$\rightarrow$2-way class} \\
   & Acc. & mAP & Acc. & AUC \\ \hline\hline
   X3D-M~\cite{feichtenhofer2020x3d}& 76.91& 95.06 & 87.51 & 93.81 \\
   SlowFast~\cite{feichtenhofer2019slowfast} & 80.86 & 95.92& 89.45 & 94.25 \\ \hlinew{1.1pt}
   \end{tabular}
   \vspace{-0.5cm}
\end{table}


\subsection{Video Forgery Classification}




In this section, we select several typical video backbones of different sizes: X3D-M~\cite{feichtenhofer2020x3d}, Slow-only R-50~\cite{feichtenhofer2019slowfast}, TSM~\cite{lin2019tsm}, and SlowFast R-50~\cite{feichtenhofer2019slowfast}. We sample 16 frames with temporal stride 4 as input to all models.

Binary classfication results of video-level forensics methods are listed in Tab.~\ref{tab:Video cls Results}.
%
Compared to image-level evaluation, video-level Acc and AUC are generally higher.
%
%
SlowFast~\cite{feichtenhofer2019slowfast} obtains the best performance on video classification, while X3D-M~\cite{feichtenhofer2020x3d}, with only a very small number of parameters, also gives satisfying results.
We select these two as representatives of large and small models respectively in subsequent experiments, as displayed in Tab.~\ref{tab:video_multiclass} and Tab.~\ref{tab:video-cross-method}.
Cross-forgery evaluation results are worse than their image counterparts, suggesting harder generalization with temporal information.

\subsection{Temporal Forgery Localization}
\label{subsec:forgery_temporal_localization}

\begin{table}[t]
   \caption{\textbf{Video Forgery Classification (Protocol 2):} binary classification. Forensics methods trained with ID-replaced forgery approaches have substantial performance drops (even more significant than their image-level counterparts) when tested on unseen ID-remained forgery approaches, and \emph{vice versa}.}
   \vspace{-0.3cm}
   \footnotesize
   \centering
   \label{tab:video-cross-method}
   \begin{tabular}{ll|rrrr}
      \hlinew{1.1pt}
      \multicolumn{2}{c|}{\multirow{2}{*}{}} & \multicolumn{2}{c}{ID-replaced} & \multicolumn{2}{c}{ID-remained} \\
      \multicolumn{2}{c|}{} & Acc. & AUC & Acc. & AUC \\ \hline\hline
      \multirow{2}{*}{X3D-M}&ID-replaced& 87.92 & 92.91 & 55.25 &65.59\\
      &ID-remained& 55.93 & 62.87 & 88.85 &95.40\\ \hline
      \multirow{2}{*}{SlowFast}&ID-replaced& 88.26 & 92.88 & 52.64 &64.83\\
      &ID-remained& 52.70 & 61.50 & 87.96 &95.47\\ \hlinew{1.1pt}
   \end{tabular}
\end{table}

We experiment with both frame-based and video-based models for temporal localization. 
For frame-based model, after binarizing frame predictions with a fixed threshold (0.25), we select consecutive fake sequences, with different tolerance levels for real frames in the middle, as final proposals. 
The confidence of a proposal is simply the average of the original frame scores.
We adopt Boundary-Sensitive Network (BSN)~\cite{lin2018bsn} and Boundary-Matching Network (BMN)~\cite{lin2019bmn} on top of X3D-M and SlowFast features as the video-based models.

Tab.~\ref{tab:video temporal localize result table} compares these methods on the validation set.
In particular, video-based methods perform significantly better than the frame-based method, demonstrating the importance of applying a boundary-aware network.
Additionally, BMN outperforms BSN with large margins, and achieves $\sim$87 average AP. This is of great significance since it shows our model is capable of effectively locating manipulated media in a large video database.
We hope our results can inspire more future works on forgery localization.


\begin{table}[]
    \scriptsize
    \centering
    \caption{\textbf{Temporal Forgery Localization.} We show AP, AR and mAP scores of all compared methods.}
    \vspace{-0.3cm}
    \label{tab:video temporal localize result table}
    \begin{tabular}{l|rr|rrr|r}
    \hlinew{1.1pt}
    \multirow{2}{*}{} & \multicolumn{2}{c|}{AR} & \multicolumn{3}{c|}{AP} &  \multirow{2}{*}{\makecell{avg.\\AP}} \\
     & 2 & 5 & 0.5 & 0.75 & 0.9 &\\ \hline
    Xception~\cite{chollet2017xception} & 25.83 & 73.95 & 68.29 & 62.84 & 58.30 & 62.83\\
    X3D-M+BSN~\cite{lin2018bsn} & 81.33 & 86.88 & 80.46 & 77.24 & 55.09 &70.29\\
    X3D-M+BMN~\cite{lin2019bmn} & 88.44&91.99&90.65 & 88.12 & 74.95 &83.47\\
    SlowFast+BSN~\cite{lin2018bsn} & 83.63 & 88.78 & 82.25 & 80.11 & 60.66 & 73.42\\
    SlowFast+BMN~\cite{lin2019bmn} & 90.64 & 93.49 & 92.76 & 91.00 & 80.02 & 86.85\\ \hlinew{1.1pt}
    \end{tabular}
    \vspace{-0.5cm}
\end{table}

\vspace{-0.2cm}
\section{Conclusion}
\label{sec:conclusion}

In this paper, we present ForgeryNet, a new mega-scale benchmark for both image- and video-level face forgery analysis. 
Compared with existing datasets for face forgery, ForgeryNet possesses more variety and is more comprehensive in terms of wild sources, various manipulation approaches, diverse re-rendering process and richness of annotations.  
We further introduce four possible applications with ForgeryNet: image and video classification, spatial and temporal localization. The results obtained in these tasks help us better understand facial forgery towards real-world scenarios.
For future works, we welcome interested researchers to contribute more novel facial forgery approaches. 
More forgery analysis can also be studied on our dataset to improve the defense capabilities.

\vspace{-0.3cm}
\paragraph{Acknowledgments} 
This work is supported by key research and development program of Guangdong Province, China, under grant 2019B010154003, as well as NTU NAP and A*STAR through the Industry Alignment Fund - Industry Collaboration Projects Grant, the National Natural Science Foundation of China under Grant No. 61906012.

{\small
\bibliographystyle{ieee_fullname}
\bibliography{egbib}
}

\appendix
\section*{Appendix}


\section{Original Data Collection}
\label{sec:original_data}

In contrast to previous facial forgery datasets~\cite{rossler2019faceforensics++,jiang2020deeperforensics} which only involve original data taken from certain briefing scenarios or TV shows, we choose four face datasets~\cite{cao2014crema,livingstone2018ryerson,Chung18b,ephrat2018looking} as the original data with diversified face identities, angles, expressions, actions, \etc, for the sake of building a wild and diverse forgery dataset. 
%

\noindent (1) \textit{CREMA-D}~\cite{cao2014crema} is a dataset of $7,442$ video clips from $48$ male and $43$ female actors with a variety of ethnicities, ages ranging from 20 to 74, and six different emotions.

\noindent (2) \textit{RAVDESS}~\cite{livingstone2018ryerson} consists of $7,356$ files including both video footages and sound tracks from $24$ professional actors with eight emotions, vocalizing two lexically-matched statements in a neutral North American accent. 

\noindent (3) \textit{VoxCeleb2}~\cite{Chung18b} is constructed with over one million YouTube videos with  utterances of $6,112$ celebrities.

\noindent (4) \textit{AVSpeech}~\cite{ephrat2018looking} is a dataset of $290$k YouTube video clips of $3\sim 10$ seconds long.
Note that the speakers talk with no audio background interference, \ie the only audible sound in the soundtrack of a video belongs to a single visible and speaking person.

\section{Original Data Preprocessing}
\label{sec:pre_process}

The selected in-the-wild videos vary in length ($2\ \text{seconds}\sim 1\ \text{hour}$), FPS ($20 \sim 30$), semantic annotations, and number of faces appearing in one frame. 
For further manipulation, we preprocess the original data into a controllable source video set:

\vspace{0.1cm}
\noindent (1) \textit{Video-Origin \& Image-Origin:}
%
Due to the large amount of videos in VoxCeleb2 and AVSpeech, we respectively pick $43,941$ and $43,584$ videos with length over 6 seconds. The videos are chosen randomly, yet in VoxCeleb2 we guarantee all $6,112$ identities are included in the selected video set.
All the selected videos from these two datasets are then truncated into $6\sim 10$ seconds to enrich length variations, while those from CREMA-D and RAVDESS are retained without cropping due to their short duration ($2\sim 5\ \text{seconds}$).
The images of \texttt{image-origin} are extracted from the aforementioned \texttt{video-origin} set with $20$ FPS.
%
%
%


\vspace{0.1cm}
\noindent (2) \textit{Target Face:}
We detect faces from images in \texttt{image-origin} by RetinaFace~\cite{Deng2020CVPR} for future manipulation. 
As shown in Fig.~\ref{fig:sourcedata} in the main paper, 
in some scenarios, multiple faces co-occur in a single frame, such as ``conversation between two or more people'' or ``crowd gathering''. 
To determine the target face for forgery, we first use a simple IoU (Intersection-over-Union) based tracking to acquire face tubes each with faces of the same person identity. We select the face which appears most frequently in the video, \ie has the longest face tube.

\vspace{0.1cm}
\noindent (3) \textit{Attribute Prediction:}
To manipulate facial attributes, the deep models require attribute labels as a conditional input. However, data in \texttt{video/image-origin} lack attribute labels due to limited annotations (\eg only ``emotions'' and ``age'') of the original datasets.
To this end, we predict the attribute labels with Slim-CNN~\cite{AttributePrediction,liu2015deep}, 
a state-of-the-art face attribute classification method.
\section{Forgery Approach}
\label{sec:forgery_approach}

To guarantee the diversity of forgery approaches in the proposed \name{}, we introduce $15$ face forgery approaches~\cite{Siarohin_2019_NeurIPS,chen2019hierarchical,fried2019text , choi2020stargan,Karras2019stylegan2 , CelebAMask-HQ, Jo_2019_ICCV,deng2020disentangled ,nirkin2019fsgan ,petrov2020deepfacelab , li2019faceshifter}, which are shown in the main paper. 
%
%
%
\begin{table*}[]
    \caption{\textbf{Summary of the four types of forgery approaches.} In this table, the input, output, architecture, resolution, modification ability, and whether to retrain in inference of each forgery approach are presented. S/T represents the modality of $x_s$ and $x_t$. v:=video, i:=image, a:=audio, m:= mask, s:=sketch, l:= noise, S:=single identity, M:=multiple identity}
    \vspace{0.1cm}
    \small
    \centering
    \label{table:forgery_approach_summary}
    \begin{tabular}{c|c|cccccc}
    \hlinew{1.1pt}
    \multicolumn{1}{l|}{} & Method & S/T & CG/GAN & Input & Modification & Resolution & Retraining \\ \hline\hline
    \multirow{3}{*}{\begin{tabular}[c]{@{}c@{}}Face\\ Reenactment\end{tabular}} & FirstOrderMotion\cite{Siarohin_2019_NeurIPS} & v/i & GAN & M/M & pose,expression & 256*256 & No need \\
     & ATVG-Net\cite{chen2019hierarchical} & v/i & GAN & M/M & pose,expression & 128*128 & No need \\
     & Talking-head Video\cite{fried2019text} & a/v & CG+GAN & M/S & mouth & 256*256 & 1$\sim$3 portraits \\ \hline
    \multirow{5}{*}{\begin{tabular}[c]{@{}c@{}}Face\\ Editing\end{tabular}} & StarGAN2\cite{choi2020stargan} & i/i & GAN & M/M & attribute transfer & 256*256 & portraits \\
     & StyleGAN2\cite{Karras2019stylegan2} & l/i & GAN & M/M & rebuild from latent & 1024*1024 & portraits \\
     & MaskGAN\cite{CelebAMask-HQ} & m,i/i & GAN & M/M & editing record & 512*512 & portraits,mask \\
     & SC-FEGAN\cite{Jo_2019_ICCV} & s,i/i & GAN & M/M & sketch record & 512*512 & portraits,sketch \\
     & DiscoFaceGAN\cite{deng2020disentangled} & i/i & CG+GAN & M/M & 3dmm attributes & 1024*1024 & portraits \\ 
    \hline
    \multirow{2}{*}{\begin{tabular}[c]{@{}c@{}}Face\\ Transfer\end{tabular}} & BlendFace & v/v & CG & M/M & identity, expression & Any & No need \\
     & MMReplacement & i/i & CG & M/M & identity, expression & Any & at least 1 protrait \\ 
     \hline
    \multirow{3}{*}{\begin{tabular}[c]{@{}c@{}}Face Swap\\ \end{tabular}} & FSGAN\cite{nirkin2019fsgan} & v/v & GAN & M/M & identity & 256*256 & No need \\
     & DeepFakes\cite{petrov2020deepfacelab} & v/v & GAN & S/S & identity & 192*192 & 2k$\sim$5k portraits \\
     & FaceShifter\cite{li2019faceshifter} & i/i & GAN & M/M & identity & 256*256 & No need \\ 
    \hlinew{1.1pt}
    \end{tabular}
\end{table*}
%
We conclude five architecture variants as, 
1) \textit{Encoder-Decoder}~\cite{faceswap} is used to disentangle the identity from identity-agnostic attributes and then modify/swap the encodings of the target before passing them through the decoder.
2) \textit{Vanilla GAN}~\cite{shen2018facefeat} consists of a generator and a discriminator which work against each other. After training, the discriminator is discarded and the generator is used to generate content.
3) \textit{Pix2Pix}~\cite{li2019faceshifter} is a popular improvement on GANs which enables translations from one image domain to another. The generator is an encoder-decoder network with skip connections from encoder to decoder which enable the generator to produce high fidelity imagery by bypassing some compression layers when needed.
In addition to the above three variants, which are the basic elements for generating a forgery image, some sequential and dynamic-length data (\eg~audio and video) are often handled by 4) \textit{RNN/LSTM} \cite{chen2019hierarchical}, and 5) \textit{Graphics Formation}~\cite{egger20203d}. The latter represents a simulation of the classical image formation process of computer graphics, that is, reconstructing a 3D face model with 3DMM parameters, which are the output of a classical analysis-by-synthesis algorithm, and then rendering the generated 3D face model into a 2D image.


\section{Re-rendering Process}
\label{sec:re_rendering}

\noindent (1) For the \textit{face mask} condition shown in Fig.~\ref{fig:pipeline} (e-1) in the main paper, we first align the landmarks of $\mathbf{\tilde{I}}_t^f$ and $\mathbf{I}_t^f$ to align their masks $\mathbf{\tilde{I}}_t^m$ and $\mathbf{I}_t^m$, and then calculate an optimal transformation to align $\mathbf{\tilde{I}}_t^f$ back to the $\mathbf{I}_t$. 
Color matching is then operated on the re-aligned face to make $\mathbf{\tilde{I}}_t^f$ more adaptable to $\mathbf{I}_t^f$\footnote{\textit{Identity-remained} forgery do not have this step since it only changes local intrinsic or external attributes. Moreover, some editing even aims at altering colors such as lip or eye color.}.
The following step is blending, with the objective of making $\mathbf{\tilde{I}}_t^f$ seamlessly fit the target full image $\mathbf{I}_t$. We corrode and blur the smaller mask between $\mathbf{\tilde{I}}_t^m$ and $\mathbf{I}_t^m$, and perform the Poisson blending along the outer contour of $\mathbf{\tilde{I}}_t^f$ to get the full forgery image $\mathbf{\tilde{I}}_t$.

\noindent (2) For the \textit{face bounding-box} condition, an easy way is to directly substitute the bounding-box in the original target image $\mathbf{I}_t^b$ with a forgery one $\mathbf{\tilde{I}}_t^b$, and simply perform the Poisson blending along the edge of the bounding-box as shown in Fig.~\ref{fig:pipeline} (e-2) in the main paper. 
However, some GAN-based approaches always induce some unexpected details outside the face region, especially some background clutters with jittery and blurred information. Meanwhile, some graphic-based approaches cannot infer the texture of non-face regions such as hair. To this end, we first calculate the convex hull of the face area through the face landmarks to obtain the face mask $\mathbf{\tilde{I}}_t^m$, and then turn to the re-rendering solution for the \textit{face mask} condition described above, as is illustrated in Fig.~\ref{fig:pipeline} (e-3) in the main paper.

Each frame of a video is re-rendered through the aforementioned steps. However, the obtained re-rendered frame sequence often contains frequent jitters due to misalignment and forgery effect. To generate a realistic and smooth video, we apply slight motion blur as well as compression or super-resolution to the frame sequence.
\section{Perturbation}
\label{sec:perturbation}
\begin{figure}[t]
  \begin{center}
    \includegraphics[width=1.0\linewidth]{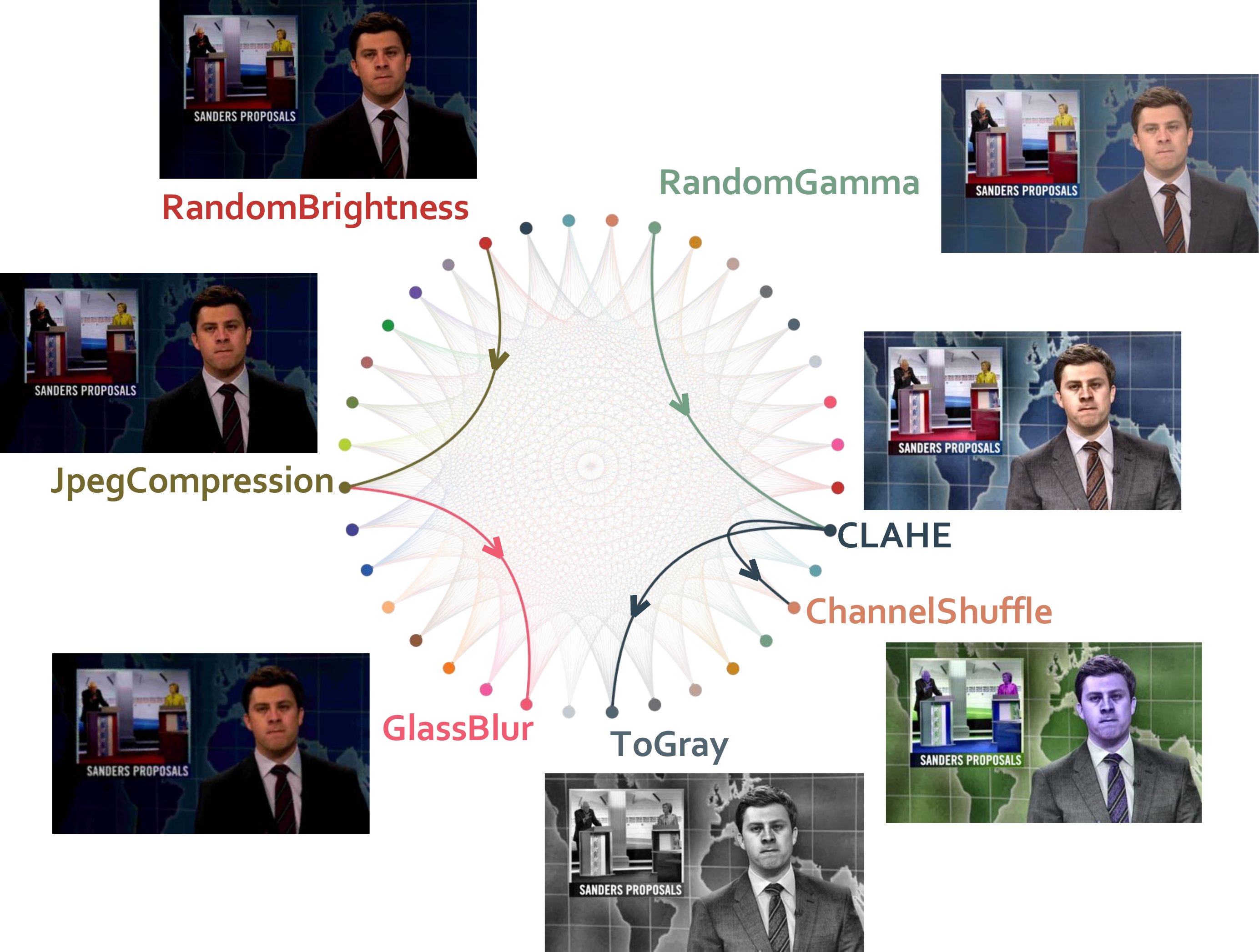}
  \end{center}
  \caption{\textbf{Perturbations in ForgeryNet.} Different perturbations are marked in different colors. This example shows the effects of one or mixed perturbations. Arrows indicate the mixture order. The image on the left is first added ``GlassBlur'' followed by ``JpegCompression'' and at last ``RandomBrightness''.}
\label{fig:augmentations}
\end{figure}

%
Fig.~\ref{fig:augmentations} presents an overview of perturbations. For example, ``GlassBlur'' and ``JpegCompression'' can simulate distortion of information in video capture and storage in the real world. Some color distortions such as ``RandomBrightness'' and ``ChannelShuffle'' provide diversity in color distributions to adapt to different color renderings of a video.

Mixed perturbations with $2\sim 4$ distortions are randomly applied to approximately $98\%$ data, while another $1\%$ are added with a single perturbation. The rest $1\%$ are remained unchanged.
Each perturbation has $1\sim 5$ intensity levels. Types and levels of the applied perturbations are all chosen at random, and are applied at the video level, \ie all frames of a video share the same type of perturbation with the same level. Meanwhile, to avoid severe distribution bias, we guarantee each pair of perturbation types co-occurs at least once.
The variety of perturbations improves the diversity and realness of ForgeryNet to better imitate the data distribution in real-world scenarios.

\section{ForgeryNet Annotation}
\label{sec:annotation}

\noindent \textbf{Image Forgery Classification.}
The annotations for this task have been elaborated in Sec.~\ref{subsec:dataset_annotation} in the main paper, where we introduce three types of forgery labels, \ie labels for two-way (real / fake), three-way (real / fake with identity-replaced forgery approaches / fake with identity-remained forgery approaches), and $n$-way ($n=16$, real and $15$ respective forgery approaches) classification tasks respectively.

\noindent \textbf{Spatial Forgery Localization.}
Due to the fact that forgery images contain various numbers of faces and each face can be manipulated completely or partially, it is more substantial to specify the manipulated area in addition to the classification labels.
We convert the forgery image $\mathbf{\tilde{I}}_t$ and the corresponding real image $\mathbf{I}_t$ into two gray-scale images to calculate their pixel-by-pixel absolute differences. We then normalize the difference map within the face area of the real image $\mathbf{I}_t^f$ to obtain a \textit{forgery distribution} $\mathbf{\tilde{I}}_t^d$. As shown in Fig.~\ref{fig:facemanipulationmask} (a) in the main paper, stronger response suggests the area is manipulated with heavier intensity.
Note that we perform perturbations on the forgery image which cause further modifications in the whole image. The perturbed forgery area distributes all over the whole image rather than merely the face region. 
In the main paper, compared to Fig.~\ref{fig:facemanipulationmask} (b) which shows a near-uniform distribution of forgery area both inside and outside the faces, the distribution before perturbation in Fig.~\ref{fig:facemanipulationmask} (a) shows its advantages in two aspects: 1) the forgery area focuses more on face area, which is consistent with how these deep forgery techniques actually work, and 2) the forgery distribution behaves distinctive among different types of forgery approaches. Take \textit{face reenactment} and \textit{face transfer} as an example, the former has particularly high response on lip and also some medium response around head since the audio- or video-source always drives the lip and pose of the target being manipulated, while the latter replaces both identity-aware and identity-agnostic contents of the target and leads to more even response inside the face.
In this paper, we define the \textit{spatial forgery localization} task as ``\textit{localizing the face area manipulated by deep forgery approaches}'', and thus the forgery distribution before perturbation $\mathbf{\tilde{I}}_t^d$ is taken as the ground-truth annotation.

\noindent \textbf{Video Forgery Classification \& Temporal Forgery Localization.}
As is mentioned in Sec.~\ref{subsec:dataset_annotation} in the main paper, in contrast to all existing datasets, we construct our video forgery dataset with untrimmed forgery videos $\mathbf{\tilde{V}}_t^{\prime}$, each of which splices real and manipulated segments together. This is based on the consideration that forgery videos in the real world often only involve manipulation on a certain subject at some key frames. 
%
Specifically, for each pair of forgery video $\mathbf{\tilde{V}}_t$ and its corresponding real video $\mathbf{V}_t$, we first randomly select $1\sim4$ segments from the forgery video $\mathbf{\tilde{V}}_t$, and then fill the rest with the corresponding real segments $\mathbf{V}_t$. Each forgery/real segment in $\mathbf{\tilde{V}}_t^{\prime}$ has no fewer than $9$ frames. 
%

Same as image-forgery, the \textit{Video Forgery Classification} also contains three types of class annotations. 
We also provide the annotations of each fragment in the untrimmed forgery video and propose a new task, \ie~\textit{Temporal Forgery Localization}, to localize the temporal segments which are manipulated. 
\section{ForgeryNet Split}
\label{sec:split}

We first split the identities of the original videos into two subsets, training and evaluation, roughly according to a proportion of 7:3. This guarantees that any person appearing in a training video does not show up in the evaluation set. Note that the AVSpeech dataset does not provide annotations on person identity, so we have to assume that different videos contain different people, and directly split the videos. The evaluation subset is then further divided into validation and test with an approximate ratio of 1:2, and there may be some identity overlaps between the validation and test subsets. The real data for our image set is sampled from the frames extracted with these original videos according to some fixed proportion. Finally, we apply our 15 forgery approaches to generate manipulated data within each subset respectively, \eg the sources and targets for generating validation forgery data must all come from the validation subset of the original videos.

\section{Image Forgery Analysis Benchmark}
\label{supp:sec:image_benchmark}

\subsection{Metrics}
\label{subsec:image_metrics}
%
\noindent\textbf{Image Forgery Classification.}
We detail calculation methods of the metrics listed in Sec.~\ref{subsubsec:Image_Forgery_Classification} in the main paper. For $k$-way classification ($k=2,\ 3,\ 16$), we use Accuracy (Acc) balanced over classes, \ie we first calculate $k$ accuracy values from the $k$ classes respectively, and then take the uniform average of them as the final balanced accuracy. We also evaluate the standard Area under ROC curve (AUC) for binary classification. In terms of the other settings with more than two classes, we turn to mean Average Precision (mAP) to measure the discrimination ability of the forensics method. More specifically, the AP of some class $i$ is simply the AUC calculated with class $i$ as the sole positive class and all others being negative. After obtaining $k$ APs, we average them to get mAP. Apart from Acc and mAP, we also compute binary metrics for 3-way or $n$-way classification, and we sum up probabilities predicted for all forgery categories as the final fake confidence.

\noindent\textbf{Spatial Forgery Localization.}
As is mentioned in Sec.~\ref{subsubsec:Image_Forgery_Localization} in the main paper, we choose three metrics for evaluating predicted maps in our spatial localization task: two variants of Intersection over Union (IoU) and L1 distance.
Let $N$ denote the number of pixels, and $\tau$ be a pre-defined threshold.
\begin{itemize}
    \item $\text{IoU} = \frac{1}{N} \sum_{i=1}^N |\mathbb{I}[\text{pred}_{i} \geq \tau] - \mathbb{I}[\text{gt}_i \geq \tau]|$ (\eg $\tau = 0.1$) represents the accuracy over all spatial grids.
    \item $\text{IoU}_\text{diff} = \frac{1}{N} \sum_{i=1}^N \mathbb{I}[|\text{pred}_{i} - \text{gt}_i| \leq \tau]$ (e.g. $\tau = 0.05$) indicates whether the predicted value of each pixel is close to the groundtruth.
    \item L1 distance $\text{Loss}_{l1} = \frac{1}{N} \sum_{i=1}^N|\text{pred}_{i} - \text{gt}_i|$ also implies how close is the predicted map to the groundtruth one.
\end{itemize}

\subsection{Models}
\label{subsec:image_models}

\noindent\textbf{Image Forgery Classification.} There are in total 11 image-level classification methods.

\begin{itemize}
    \item \textbf{MobileNetV3}~\cite{howard2019searching} is an efficient mobile model, combining the following three layers: depthwise separable convolutions from MobileNetV1~\cite{howard2017mobilenets}, the linear bottleneck and inverted residual structure from MobileNetV2~\cite{sandler2018mobilenetv2}, and lightweight attention modules based on squeeze and excitation from MnasNet~\cite{tan2019mnasnet}. We use both MobileNetV3-Small and MobileNetV3-Large for evaluation.

    \item \textbf{EfficientNet-B0}~\cite{tan2019efficientnet} is the baseline network of the EfficientNet family, which is developed by leveraging a multi-objective neural architecture search based on mobile inverted bottleneck MBConv~\cite{sandler2018mobilenetv2} with squeeze-and-excitation optimization~\cite{hu2018squeeze} added to it.

    \item \textbf{ResNet-18}~\cite{he2016deep}  is the smallest ResNet architecture with 17 convolutional layers and one fully connected layer for final output.

    \item \textbf{Xception}~\cite{chollet2017xception} is a deep convolutional network architecture based on Inception replaced with depthwise separable convolutions. Xception is regarded as our default baseline in further experiments.

    \item \textbf{ResNeSt-101}~\cite{zhang2020resnest} is a new variant of ResNet. It introduces a modular Split-Attention block that enables attention across different feature-map groups and stacks these blocks ResNet-style to get better performance with similar number of parameters.

    \item \textbf{SAN19-patchwise}~\cite{zhao2020exploring} takes patchwise self-attention as the basic building block for image recognition. Specifically, we uses SAN19 which roughly corresponds to ResNet-50 to evaluate.

    \item \textbf{ELA-Xception} and \textbf{SNRFilters-Xception} differ from Xception in the fact that they do not directly take RGB images as input. More specifically, the input for ELA-Xception is the resulting difference image from Error Level Analysis (ELA)~\cite{gunawan2017development}. SNRFilters-Xception, as its name suggests, applies a set of $5\times5$ high pass kernels~\cite{chen2017jpeg} to the original input image, and then concatenate the 4 output images along the channel dimension (the number of input channels of the first convolution in Xception is changed to 12 accordingly).

    \item \textbf{Gram-Net} designs Gram Block to leverage global image texture information for fake image detection. The original paper~\cite{liu2020global} adds Gram Blocks to the ResNet architecture. Yet in our benchmark, we apply them to our baseline model Xception for the sake of fair comparison.

    \item \textbf{F$^\mathbf{3}$-Net}~\cite{qian2020thinking} explores frequency information for face forgery detection by taking advantages of two frequency-aware clues: frequency-aware decomposed image components and local frequency statistics. Note that F$^3$-Net also uses Xception as the backbone network.
\end{itemize}

\noindent\textbf{Spatial Forgery Localization.} We select 3 representative models for spatial localization.

\begin{itemize}
    \item \textbf{Xception+Regression} uses Xception as the backbone network, and adds an extra deconvolution layer after the final feature map to form a direct regression branch which outputs the spatial forgery map.
    
    \item \textbf{Xception+UNet}~\cite{ronneberger2015u} supplements a usual contracting network by successive layers where pooling operations are replaced by upsampling operators. A successive convolutional layer can learn to assemble a precise output based on this information. For fair comparison, UNet also uses Xception as its encoder network. 
    
    \item \textbf{HRNet}~\cite{wang2020deep} starts from a high-resolution convolution stream, gradually adds high-to-low resolution convolution streams, and connects the multi-resolution streams in parallel. We use the HRNet-W48 instantiation.
\end{itemize}

\subsection{Implementation Details}
\label{subsec:image_impl}

\noindent\textbf{Training.}
For classification methods, we use the default cross-entropy loss for training. As for localization methods, we also add a segmentation loss in addition to the classification loss. There are two choices for the segmentation loss: (1) binary cross entropy loss with soft targets averaged over all spatial locations; (2) MSE loss with respect to groundtruth targets. We select one of these two losses for each localization model based on validation results.

All models use ImageNet~\cite{deng2009imagenet} for pre-training. We train both classification and localization models end-to-end using synchronous SGD for optimization. The mini-batch size is set to 128. We adopt a multistep learning rate schedule with 100k iterations in total, and the learning rate is decreased by a factor of 0.5 at steps 20k, 40k, 60k, 70k, 80k and 90k. The base learning rate for each model is selected from the set $\{0.01, 0.02, 0.05\}$ according to validation performance. We use linear warm-up~\cite{goyal2017accurate} from 0.01 during the first 1k iterations. The weight decay is set to $10^{-4}$ and we apply Nesterov momentum of 0.9. We use face images cropped with provided square bounding boxes (detected boxes enlarged $1.3\times$) for training. For data augmentation, we with 99\% probability randomly select one perturbation from some set of perturbation methods, and apply it to the input image. Apart from random perturbation, for a model with input spatial size $S\times S$, we scale the side length to a random value in range $[S, 8S/7]$, and then randomly crop out a $S\times S$ region. Note that for five Xception-based classification models $S=299$, for three localization models $S=256$, and for the other six classification models $S=224$. We also apply random horizontal flip before feeding the input to the model.

\noindent\textbf{Inference.}
We only perform single-crop inference, and directly scale the input face image to the input spatial size $S\times S$ of the model.

\subsection{More Experiments}
\label{subsec:image_more_exps}

\noindent\textbf{Ablation Study on Augmentation.}
We experiment on three different levels of augmentation: weak, normal and enhanced. Weak augmentation does not add random perturbation mentioned in Appendix~\ref{subsec:image_impl}, while normal and enhanced settings include different numbers of common perturbation methods in the perturbation set for augmentation. Results of Xception trained on these types of data augmentation are shown in Tab.~\ref{tab:aug ablation study}. It can be seen that exerting appropriate augmentation to the training set significantly improves the performance of an image forgery classification model.
\begin{table}[]
   \caption{\textbf{Ablation study on augmentation (image).} We report accuracy and AUC scores of Protocol 1 binary classification on the validation set with three different levels of augmentation.}
   \vspace{-0.3cm}
   \small
   \centering
   \label{tab:aug ablation study}
   \begin{tabular}{l|rr|rr|rr}
   \hlinew{1.1pt}
   \multirow{2}{*}{} & \multicolumn{2}{c|}{weak aug} & \multicolumn{2}{c|}{normal aug} & \multicolumn{2}{c}{enhanced aug}  \\
     & Acc. & AUC & Acc. & AUC & Acc. & AUC \\ \hline\hline
   Xception  & 66.73 & 74.75 & 73.70 & 82.56 & 80.78 & 90.12  \\
 \hlinew{1.05pt}
   \end{tabular}
   \vspace{-0.1cm}
\end{table}

\noindent\textbf{Cross-dataset Experiments.}
We provide cross-dataset testing results with our \name{} (image forgery binary classfication only) as well as three public deepfake datasets - FF++ (c23)~\cite{rossler2019faceforensics++}, DFDC~\cite{dolhansky2020deepfake}, and DeeperForensics-1.0 (DF1.0)~\cite{jiang2020deeperforensics} which are only used for testing. For evaluation, we use (1) test set of FF++ (c23); (2) both validation and test set (only the released half) of DFDC; (3) a subset of DF1.0 which corresponds to the test set of FF++; (4) test set of our image benchmark. For video datasets, we extract frames with temporal stride 30 for frame-level testing. We present the numbers in Tab.~\ref{tab:cross-dataset}. \name{} shows the best cross-dataset performances on all other test sets, which indicates the strong generality of our dataset.
\begin{table}[]
\scriptsize
\centering
\caption{\textbf{Cross-dataset experiments.} We report frame-level AUC scores. Each row corresponds to a model trained with one of the datasets. Underlined values are results of models trained and tested on the same dataset, and the bold ones emphasize best cross-dataset performances.}
\vspace{-0.3cm}
\label{tab:cross-dataset}
\begin{tabular}{P{1.4cm} | M{0.6cm} M{0.6cm} M{1.1cm} M{1.1cm} M{1.0cm}}
\hlinew{1.1pt}
 & DF1.0 & FF++ & DFDC(val) & DFDC(test) & ForgeryNet \\ \hline\hline
FF++~\cite{rossler2019faceforensics++} & 85.41 &  \underline{99.43} & 59.77 & 62.19 & 63.80 \\
DFDC~\cite{dolhansky2020deepfake} & 79.60 & 71.34 &  \underline{90.12} &  \underline{93.50} & \textbf{68.93} \\
\name{} & \textbf{90.09} & \textbf{85.06} & \textbf{69.68} & \textbf{71.08} & \underline{90.09} \\ \hlinew{1.1pt}
\end{tabular}
\end{table}

\section{Video Forgery Analysis Benchmark}
\label{supp:sec:video_benchmark}

\subsection{Metrics}
\label{subsec:video_metrics}
\noindent\textbf{Video Forgery Classification.}
The metrics for this task are the same as those for image classification.

\noindent\textbf{Temporal Forgery Localization.}
For the temporal localization task, the goal is to generate proposals which have high temporal overlap with the groundtruth (manipulated segments) as well as high recall.
We give specifics on our employed metrics for evaluating predicted segments with respect to the groundtruth ones, which are Average Precision at some tIoU threshold (AP@$t$, \eg $t=0.5$), average AP, as well as Average Recall@$K$ (AR@$K$, \eg $K=5$). Note that these metrics mostly reference ActivityNet~\cite{ghanem2018activitynet} evaluation.
%
In details, we choose 10 equally-spaced tIoU threshold values between 0.5 and 0.95 (inclusive) with a step size of 0.05. Under a certain tIoU threshold value $t$, we may match our predicted segments with the groundtruth according to the condition that tIoU $\geq t$. Recall@$K$ with tIoU threshold $t$ is defined as the proportion of groundtruth which can be matched with some prediction, after preserving only $K$ predicted segments per video on average. AP@$t$, on the other hand, is the Area under ROC curve computed with predictions and their associated confidence scores, treating the predictions which are matched to some groundtruth segment with tIoU threshold $t$ as positive. Finally, average AP and AR@$K$ are simply the uniform average of APs and Recall@$K$s computed at the 10 tIoU thresholds, respectively. Note that both real and fake videos are included in our evaluation, although the real ones do not contain any forgery segment (Recall is not be affected by real videos, but AP is).

\subsection{Models}
\label{subsec:video_models}

\noindent\textbf{Video Forgery Classification.}  We choose four typical models for video classification.

\begin{itemize}
    \item \textbf{TSM}~\cite{lin2019tsm} inserts Temporal Shift Modules to 2D CNNs to achieve temporal modeling at zero computation and zero parameters. We follow its default setting with ResNet-50 as the backbone network.
    
    \item \textbf{SlowFast}~\cite{feichtenhofer2019slowfast}, featuring its two-pathway design with different input temporal strides, is one of the state-of-the-art architectures for action recognition. We choose its R-50 instantiation (without Non-Local blocks), and set the fast-to-slow ratio $\alpha=4$.
    
    \item \textbf{Slow-only} is basically the slow pathway of SlowFast, and we also use the R-50 instantiation. Note that with the same number of input frames, Slow-only is actually heavier than SlowFast since the slow branch of the latter only use $1/\alpha$ of the frames.
    
    \item \textbf{X3D-M}~\cite{feichtenhofer2020x3d} is one member of the X3D family, a series of efficient video networks obtained by progressive expansion along multiple axes. It is able to achieve performances nearly comparable with SlowFast R-50 on common video benchmarks while having much fewer parameters.
\end{itemize}

\noindent\textbf{Temporal Forgery Localization.}
As described in Sec.~\ref{subsec:forgery_temporal_localization} in the main paper, we include a frame-based method, where we use Xception as the frame prediction model. The logic of this method can be briefly stated as the following:
\begin{enumerate}
    \item For a video with $T$ frames, we run the Xception model to get frame-level scores, and then binarize them with threshold 0.25, acquiring a sequence of $T$ binary predictions (real/fake).
    \item We enumerate tolerance value in the set $\{1,3,5,7\}$. For a tolerance value $t$, we inspect the sequence of $T$ predictions, and selects manipulated segments with at least 5 frames satisfying that the length of consecutive real frames in the middle does not exceed $t$. The confidence score of a segment is simply the average of its frame-level scores.
    \item We combine segments predicted with different tolerance levels, and remove duplicates to form the final predictions.
\end{enumerate}

For two video-based methods (BSN~\cite{lin2018bsn} and BMN~\cite{lin2019bmn}), we use SlowFast and X3D-M for extracting clip features, forming four different ``feature+method" pairs. Note that for these feature extraction models, we use fewer input frames for training than their classification counterparts to increase temporal locality. Accordingly, the fast-to-slow ratio $\alpha$ of SlowFast is decreased to $2$.

\begin{figure*}[t]
  \begin{center}
    \includegraphics[width=1.0\linewidth]{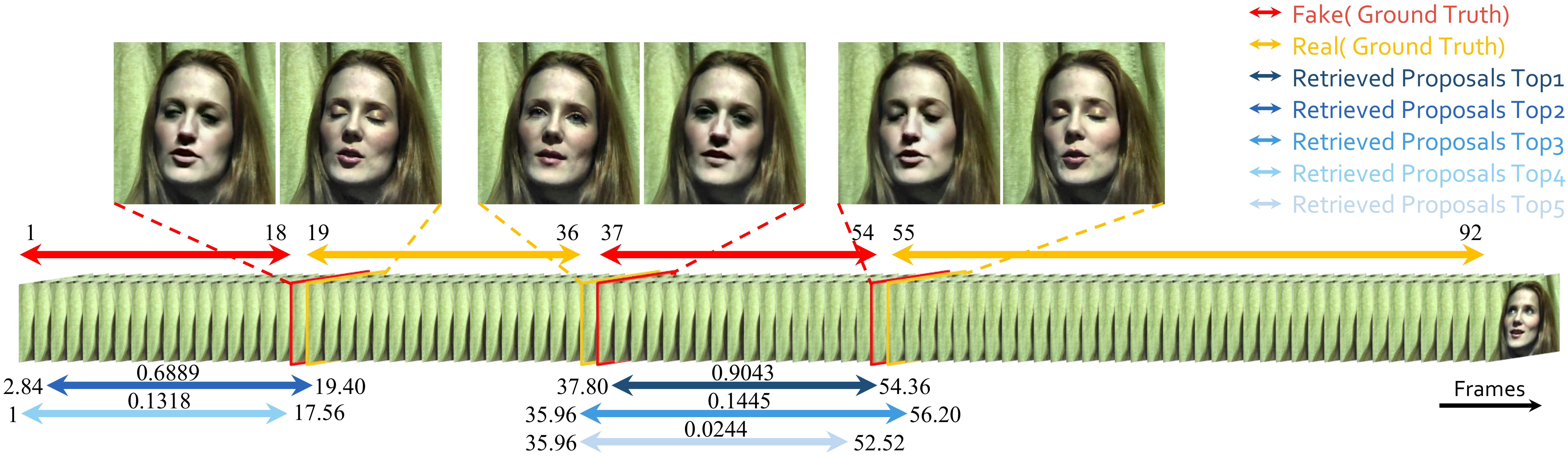}
  \end{center}
  \caption{\textbf{Example of temporal forgery localization.} We show top-5 predictions of the model SlowFast+BMN. All endpoints of the two manipulated segments are localized with high precision.}
\label{fig:VTLA}
\end{figure*}

\subsection{Implementation Details}
\label{subsec:video_impl}

\noindent\textbf{Training.}
For classification methods and feature extraction models for localization, we use the default cross-entropy loss for training. The frame-based localization method directly uses the Xception model trained with the image binary classification task, and does not need any extra training. BSN and BMN have their own training loss functions and procedures which we do not alter.

All models use Kinetics-400~\cite{carreira2017quo} for pre-training. We train them end-to-end using synchronous SGD for optimization. The mini-batch size is set to 64. We adopt a multistep learning rate schedule with 50k iterations in total, and the learning rate is decreased by a factor of 0.5 at steps 20k, 30k, 40k and 45k. The base learning rate is set to 0.02. We use linear warm-up from $10^{-3}$ during the first 500 iterations. All classification models take 16 frames with a temporal stride of 4 as input, yet the feature extraction models (SlowFast and X3D-M) for BSN and BMN use only continuous 8 frames as input for better temporal sensitivity. We use temporal random crop for training, \ie for a model requiring $T$ frames $\times$ stride $\tau$, we randomly sample a segment of length $T\times\tau$ from the video. In some rare cases where the entire video has less than $T\times\tau$ frames, we use loop padding to fill the rest. The input spatial size is fixed to $S=224$. Other training details are the same as those for image experiments.

For BSN and BMN, since the feature extraction models take 8 frames as input, we extract features with stride $4$. We set the temporal scale parameter to 50, and linearly interpolate the extracted features to the 51 endpoints. We only use fake videos for training video-based localization models. We train TEM and PEM in BSN for 20 epochs each. We train BMN for 9 or 18 epochs according to validation performance. The mini-batch size is set to 128. Other hyper-parameters follow the original settings of BSN and BMN.

\noindent\textbf{Inference.}
We scale the input to $S\times S$ spatially. On the temporal dimension, we use two settings for classification inference (suppose input temporal sampling is $T\times\tau$): (1) single-crop, or to be more specific, temporally center crop $T\times\tau$ frames; (2) multi-crop, \ie crop multiple segments of length $T\times\tau$ to cover the entire video.

For temporal localization, we only keep top $10$ predictions per video in terms of confidence score, and for video-based methods, relevant hyper-parameters are the same as training.

\subsection{More Experiments}
\label{subsec:video_more_exps}

\noindent\textbf{Ablation Study on Augmentation.}
We conduct similar experiments on augmentation with the same settings as Appendix~\ref{subsec:image_more_exps}. As presented in Tab.~\ref{tab:video_aug_ablation_study}, we observe that our video-level forgery classification method is less affected by augmentation than its image-level counterpart.
\begin{table}[]
   \caption{\textbf{Ablation study on augmentation (video).} We report accuracy and AUC scores of Protocol 1 binary classification on the validation set with three different levels of augmentation.}
   \vspace{-0.3cm}
   \small
   \centering
   \label{tab:video_aug_ablation_study}
   \begin{tabular}{l|rr|rr|rr}
   \hlinew{1.1pt}
   \multirow{2}{*}{} & \multicolumn{2}{c|}{weak aug} & \multicolumn{2}{c|}{normal aug} & \multicolumn{2}{c}{enhanced aug}  \\
     & Acc. & AUC & Acc. & AUC & Acc. & AUC \\ \hline\hline
   SlowFast  & 84.39 & 91.61 & 87.75 & 93.22 & 88.78 & 93.88  \\
 \hlinew{1.05pt}
   \end{tabular}
   \vspace{-0.1cm}
\end{table}

\noindent\textbf{Temporal Shuffling Experiments.}
To verify the effect of continuous temporal information for video forgery classification, we train the SlowFast model with different levels of temporal random shuffling to disrupt temporal continuity: shuffle every 16 frames, shuffle every 64 frames, and shuffle all frames. The results in Tab.~\ref{tab:video_shuffle_study} indicate that temporal disruptions have considerable, but not very major impact on the performance video classification, implying the video model may have leveraged other sources of information than the continuous temporal flow. An interesting finding is that a weak level of random shuffling (shuffle 16) even slightly boosts the AUC score compared to the setting without shuffling recorded in Tab.~\ref{tab:video_aug_ablation_study}.
\begin{table}[]
   \caption{\textbf{Experiemnts on temporal shuffling.} We report accuracy and AUC scores of Protocol 1 binary classification on the validation set with three different levels of temporal shuffling.}
   \vspace{-0.3cm}
   \small
   \centering
   \label{tab:video_shuffle_study}
   \begin{tabular}{l|rr|rr|rr}
   \hlinew{1.1pt}
   \multirow{2}{*}{} & \multicolumn{2}{c|}{shuffle 16} & \multicolumn{2}{c|}{shuffle 64} & \multicolumn{2}{c}{shuffle all}  \\
     & Acc. & AUC & Acc. & AUC & Acc. & AUC \\ \hline\hline
   SlowFast  & 88.63 & 94.11 & 86.24 & 93.00 & 85.04 & 91.74  \\
 \hlinew{1.05pt}
   \end{tabular}
   \vspace{-0.1cm}
\end{table}

\subsection{Temporal Localization Analysis}
\label{subsec:temporal_loc_analysis}
We present an example of temporal forgery localization in Fig.~\ref{fig:VTLA}. This data sample demonstrates the ability of a boundary-aware model to locate the transitions between real and fake. All endpoints are accurately pointed out by the BMN model. Note that there exist some highly similar predictions, yet are suppressed by a SoftNMS process.

\end{document}